\documentclass[10pt,twocolumn,letterpaper]{article}

\usepackage[pagenumbers]{cvpr} % To produce the arXiv/public version with page numbers

\usepackage{placeins}
\usepackage{cuted}
\usepackage{multirow}
\usepackage{colortbl}

\definecolor{mrvblue}{RGB}{226,238,248}
\definecolor{mrvpink}{RGB}{249,230,234}
\definecolor{mrvyellow}{RGB}{250,242,214}
\definecolor{mrvgreen}{RGB}{220,245,220}

\usepackage{fvextra}
\usepackage{tcolorbox}

\definecolor{suppSoftBlue}{RGB}{230,245,255}
\definecolor{suppBlue}{RGB}{54,125,189}
\definecolor{suppSoftGreen}{RGB}{220,245,220}
\definecolor{suppGreen}{RGB}{48,112,64}
\definecolor{suppSoftYellow}{RGB}{255,242,204}
\definecolor{suppOrange}{RGB}{166,94,22}
\definecolor{suppSoftPurple}{RGB}{245,235,255}
\definecolor{suppPurple}{RGB}{112,66,145}
\definecolor{suppSoftOrange}{RGB}{255,235,210}
\definecolor{suppDeepOrange}{RGB}{170,78,35}
\definecolor{suppSoftPink}{RGB}{255,230,240}
\definecolor{suppPink}{RGB}{162,65,101}
\definecolor{suppSoftTeal}{RGB}{226,247,245}
\definecolor{suppTeal}{RGB}{42,123,119}

\tcbset{
  suppprompt/.style={
    fonttitle=\small\bfseries,
    coltitle=white,
    boxrule=0.55pt,
    arc=1.4mm,
    outer arc=1.4mm,
    top=3pt,
    bottom=3pt,
    left=4pt,
    right=4pt,
    before skip=6pt,
    after skip=6pt
  }
}

\definecolor{cvprblue}{rgb}{0.21,0.49,0.74}
\usepackage[pagebackref,breaklinks,colorlinks,allcolors=cvprblue]{hyperref}

\makeatletter
\patchcmd{\@maketitle}{\vskip .5em}{\vskip 0pt}{}{}
\patchcmd{\@maketitle}{\vspace*{12pt}}{\vspace*{0pt}}{}{}
\makeatother

\def\paperID{*****}
\def\confName{CVPR}
\def\confYear{2026}

\title{RefCaptioner: Multi-Reference Image-Grounded Video Captioning}

\author{
    Tengfei Liu$^{1}$ \,
    Yang Shi$^{1,2}$\thanks{Corresponding Author} \,\,
    Yuran Wang$^{1}$ \,
    Xiaohan Zhang$^{3}$ \,
    Yuqing Wen$^{4}$ \,
    Yuqi Tang$^{5}$ \,
    \\
    Qixun Wang$^{1}$ \,
    Zhuoran Zhang$^{1}$ \,
    Xuanyu Zhu$^{1}$ \
    Weihong Lin$^{2}$ \
    Xinlei Yu$^{6}$ \,
    Yujie Wei$^{7}$ \,
    Xinwei Long$^{8}$ \,
    \\
    Fengxiang Wang$^{9}$ \,
    Xinlong Chen$^{10}$ \,
    Yue Ding$^{10}$ \,
    Jialu Chen$^{2}$ \,
    Haotian Wang$^{8}$\footnotemark[1] \,\,
    Yuanxing Zhang$^{2}$\thanks{Project Lead}
    \\ 
    {\small
    $^1$PKU \enspace
    $^2$Kling Team \enspace
    $^3$NUS \enspace
    $^4$NJU \enspace
    $^5$HKUST(GZ) \enspace
    $^6$CUHK \enspace
    $^7$FDU \enspace
    $^8$THU \enspace
    $^9$Shanghai AI Lab \enspace
    $^{10}$CASIA}
    \\
    {\centering}
    \url{https://github.com/pkucs-Ltf/RefCaptioner}
}

\begin{document}
\maketitle
\setlength{\stripsep}{0pt}
\begin{strip}
  \centering
  \includegraphics[width=1.0\textwidth]{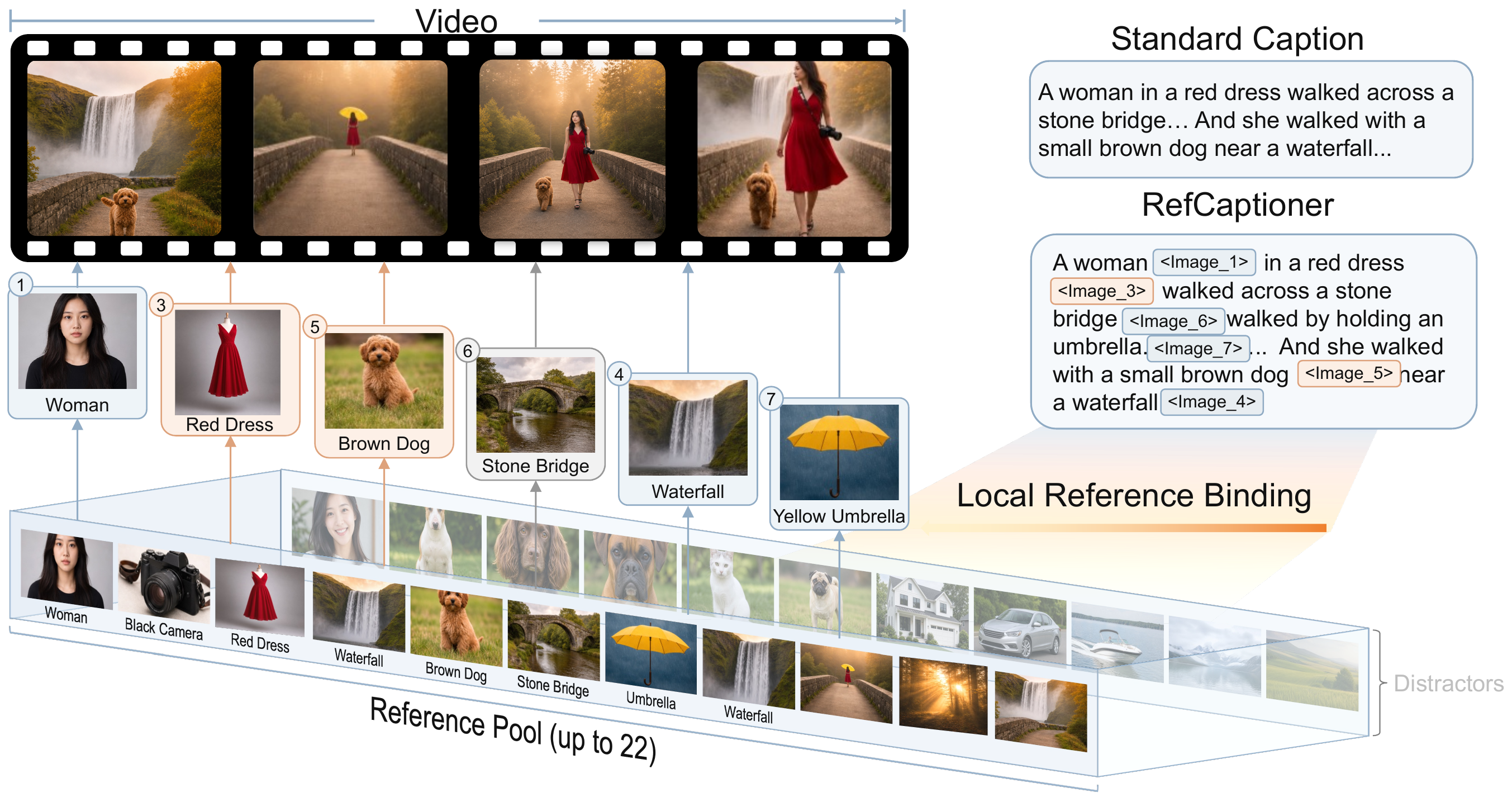}
  \captionof{figure}{\textbf{Motivation:} Video captioning models can generally describe video content clearly, but still have difficulty correctly aligning multiple reference images with the corresponding visual facts.}
  \label{fig:teaser}
  \vspace{0.4em}
\end{strip}

\begin{abstract}
Existing video captioning models generate natural descriptions of video content but cannot explicitly ground local visual elements to multiple reference images. 
We introduce multi-reference image-grounded video captioning, a new task requiring factual video descriptions with phrase-level reference grounding, and propose \textbf{RefCaptioner}, a two-stage post-training framework for this task. 
RefCaptioner combines mixed-data SFT with Hierarchical Coverage-Discounted GRPO to jointly improve reference selection, phrase-level binding, distractor rejection, and cross-reference consistency while preserving general video-captioning ability. 
To support training, we construct a corpus containing $20,000$ videos and $171,354$ reference images. 
We further introduce \textbf{MRVBench}, a benchmark for evaluating caption factuality and multi-reference grounding on both real-world and AI-generated videos. 
Experiments show that RefCaptioner achieves the best overall performance among the open-source models while remaining competitive on standard video captioning benchmarks. 
Human evaluation further confirms that its captions are preferred by annotators and enable more source-faithful video reconstruction with both open-source and proprietary video generators.
\end{abstract}
    
\section{Introduction}
\label{sec:intro}

Multimodal large language models (MLLMs) have substantially advanced video understanding, including video question answering~\cite{zhang2023videollama,maaz2024videochatgpt,shi2025mavors,zhang2025debiasing,wang2025monet,tang2026artifact}, temporal reasoning~\cite{lin2024videollava,shi2026mme,wang2024internvideo2}, and video captioning~\cite{chai2024auroracap}. 
However, most existing paradigms treat the video as a standalone input, without requiring explicit correspondence with multiple reference images. 
Joint understanding requires matching entities, appearances, and scenes in the video to relevant reference images while filtering out distractors.
This capability is fundamental to multi-reference video generation, editing, and understanding, all of which rely on accurate reference–video alignment. 
Recent studies have increasingly explored reference-conditioned video generation~\cite{chen2025videoalchemist,liu2026longav,tang2026keyframe,liang2025movieweaver,song2026mvs2v,xing2026lumosx} and editing~\cite{liu2026rebind,lin2026kiwiedit}. 
Nevertheless, comparatively little attention has been paid to explicitly expressing these fine-grained correspondences in video captioning.

We therefore introduce a new task, \emph{multi-reference image-grounded video captioning}. Given a video and several reference images, the model generates a caption and places each image tag after the visual phrase described by that reference. The caption must describe the video clearly and use the correct references. Traditional video captions~\cite{ge2025archunyuan,yang2025keyevl15} may provide either coarse or detailed descriptions, but they do not identify the relations between the video content and the reference images. For example, in Figure~\ref{fig:teaser}, the model should associate the woman, red dress, dog, umbrella, bridge, and waterfall with their corresponding reference images and reject candidates that are not supported by the video.

This task presents three challenges beyond standard video captioning. \textit{(1) Ambiguous reference selection.} The model must identify which candidate images are supported by the video, even when they are visually similar, partially relevant, redundant, or unrelated. Using every image improves coverage but introduces false references, while avoiding image tags rejects distractors at the cost of useful grounding. \textit{(2) Phrase-level reference binding.} Each selected reference must be placed after the correct local phrase. In multi-subject videos, a model may select all relevant images yet still exchange identities, attributes, or actions between subjects. \textit{(3) Cross-reference consistency.} Multiple images may depict the same subject from different viewpoints or appearances and should be grouped after the same phrase rather than treated as separate entities. The model must solve all three problems while producing a factual, detailed, and temporally coherent caption.

To address these challenges, we propose \textbf{RefCaptioner}, a two-stage post-training framework that generates factual video captions with phrase-level reference grounding. 
The first stage performs supervised fine-tuning (SFT) on a balanced mixture of manually verified multi-reference captions and detailed general video captions. 
The former teach reference selection and phrase-level tag binding, while the latter preserve factuality, coverage, and descriptive detail. 
Building on the SFT model, the second stage applies \textbf{Hierarchical Coverage-Discounted GRPO (HCD-GRPO)} to jointly optimize factual captioning and reliable reference grounding. 
Its factuality branch rewards correct and complete video descriptions, whereas its grounding branch rewards valid phrase-reference bindings, suppresses unsupported references through Distractor-Aware Evidence Suppression (DAES), and promotes consistent grouping of references depicting the same entity through Cross-Reference Semantic Coherence (CRSC).

To evaluate multi-reference video understanding, we construct \textbf{MRVBench}, comprising $462$ real-world and AI-generated videos, $3,831$ reference images, and $2,172$ question–answer pairs for factuality evaluation. 
MRVBench evaluates caption factuality, reference selection, phrase-level binding, distractor rejection, and cross-reference consistency. 
Experiments show that RefCaptioner achieves the best overall performance among the open-source models while remaining competitive on the general video captioning benchmarks VDC and VCapsBench. 
Human evaluation further validates its caption quality and phrase-level grounding accuracy. 
We also assess the downstream utility of its grounded captions through caption-conditioned video reconstruction. 

Our contributions are summarized as follows:
\begin{itemize}
    \item We propose \textbf{RefCaptioner}, a two-stage post-training framework that combines mixed-data SFT with HCD-GRPO to jointly optimize video captioning and phrase-level reference grounding, including reference selection, distractor rejection, and cross-reference consistency.

    \item We construct \textbf{MRVBench}, comprising $462$ real-world and AI-generated videos, $3,831$ reference images, and $2,172$ question--answer pairs for factuality evaluation, to assess caption factuality and multi-reference grounding.

    \item Extensive experiments demonstrate that RefCaptioner achieves the best performance among the open-source models while remaining competitive on general video captioning benchmarks. Human evaluation validates its caption quality and grounding accuracy, while caption-conditioned video reconstruction further supports the downstream utility of its generated captions.
\end{itemize}

\begin{figure*}[t]
  \centering
  \includegraphics[width=0.96\textwidth]{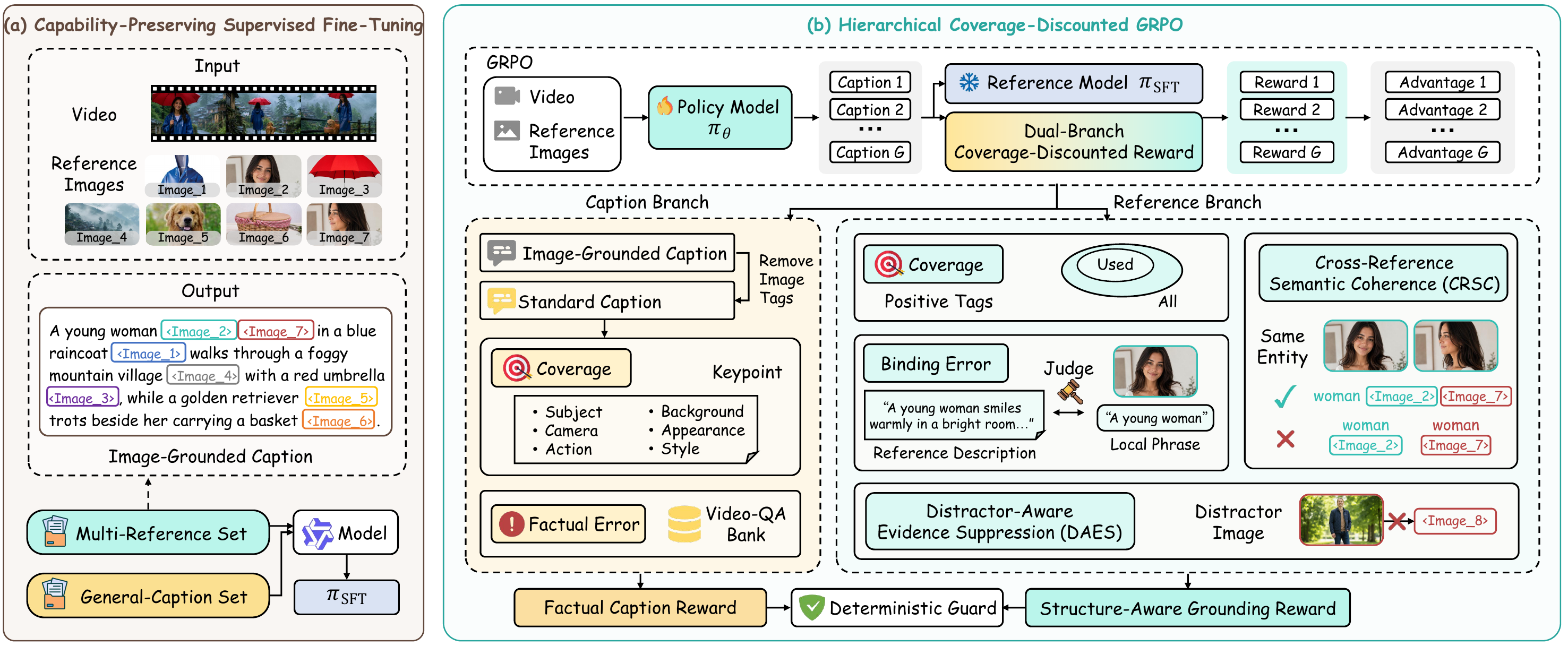}
  \caption{\textbf{Overview of our post-training framework.}
  Starting from Qwen3-VL-8B, multi-reference SFT initializes the caption format and local image-tag interface.
  GRPO then uses a layered reward evaluator to improve reference-to-visual-unit binding.}
  \label{fig:arc}
\end{figure*}
\section{Related Work}
\label{sec:related-work}

\paragraph{Video Captioning.}
Video captioning has evolved from generating short clip summaries to producing detailed descriptions with richer temporal understanding. 
Early video–language pre-training methods, such as VideoBERT and UniVL, learned shared video–text representations, while SwinBERT enabled end-to-end caption generation with video Transformers~\cite{sun2019videobert,luo2020univl,lin2022swinbert}. 
More recent multimodal models, including Tarsier and AuroraCap, generate detailed descriptions of subjects, attributes, actions, scenes, and camera changes~\cite{wang2024tarsier,chai2024auroracap}. 
Recent work has further extended captioning to long-form and fine-grained settings. 
Video ReCap and LongCaptioning generate hierarchical or long-form descriptions for extended videos, whereas ProgressCaptioner emphasizes frame-level temporal precision~\cite{islam2024videorecap,wei2025longcaptioning,xue2025progresscaptioner}. 
These advances have substantially improved the richness and temporal fidelity of video captions.

\paragraph{Video Caption Benchmarks.}
Early benchmarks such as MSR-VTT~\cite{xu2016msrvtt} and VATEX~\cite{wang2019vatex} established large-scale open-domain and multilingual testbeds, typically evaluated with lexical-overlap metrics such as BLEU and CIDEr. 
While effective for concise captions, these metrics are less suitable for long, free-form descriptions. 
Recent benchmarks therefore emphasize factual correctness and fine-grained content: DREAM-1K~\cite{wang2024tarsier} and VDC~\cite{chai2024auroracap} evaluate verifiable visual facts, while VCapsBench~\cite{zhang2026vcapsbench} and VidCapBench~\cite{chen2025vidcapbench} assess correctness, coverage, and spatial, temporal, and motion information. 
Temporal and structural consistency have also become important, as reflected in FrameCapEval~\cite{xue2025progresscaptioner}, Video ReCap~\cite{islam2024videorecap}, and CapRiCorn-1K~\cite{chen2026capricorn}. 
Meanwhile, reference-conditioned benchmarks have evolved from single-subject preservation to multi-reference composition, including OpenS2V-Nexus~\cite{yuan2025opens2vnexus} and MultiRef-Compass~\cite{zhang2026multirefcompass,deng2026magref}. 
However, none evaluates whether MLLMs can explicitly ground multiple reference images in video captions. 
We therefore introduce MRVBench for multi-reference image-grounded video captioning.

\section{Problem Formulation}
\label{sec:problem-formulation}

Multi-reference video captioning predicts observable semantic correspondence between candidate references and video content, rather than causal generation provenance.
Given a video $V$ and a set of reference images $\mathcal{I}=\{I_1,\ldots,I_N\}$, the goal is to generate a detailed and faithful caption $Y$ with explicit local grounding tags \texttt{<Image\_i>}. A valid output should satisfy three requirements. \textbf{Accuracy}: each image tag should be placed immediately after a phrase that semantically matches the referenced image. \textbf{Identification}: reference images whose content does not appear in the video should be identified as distractors and left unused. \textbf{Consistency}: when multiple reference images depict the same subject, their tags should be grouped after the same phrase. Unlike ordinary video captioning, this task requires both detailed video description and reliable phrase-level reference grounding.

\section{Method}
\label{sec:method}

\subsection{Overview}

We use Qwen3-VL-8B-Instruct as the base model for RefCaptioner and train it in two post-training stages. 
As shown in Figure~\ref{fig:arc}, the first stage performs supervised fine-tuning on a mixture of multi-reference and standard video-caption data. 
The second stage applies \textbf{Hierarchical Coverage-Discounted GRPO (HCD-GRPO)} to improve factual captioning and multi-reference grounding.

\subsection{Mixed-Data Supervised Fine-Tuning}
\label{sec:multi-reference-sft}

We construct the SFT data from two sources. 
The multi-reference set $\mathcal{D}_{\mathrm{mr}}$ contains gold-standard captions with local image tags. 
These samples are manually checked for video factuality, tag placement, and image--phrase correspondence. 
The general-caption set $\mathcal{D}_{\mathrm{cap}}$ contains detailed descriptions collected from standard video-caption training data. 
We sample the two sets at an equal ratio during training. $\mathcal{D}_{\mathrm{mr}}$ teaches the model to select relevant references and place each image tag after the corresponding caption phrase, while $\mathcal{D}_{\mathrm{cap}}$ retains its ability to describe video content in detail. 
We optimize the standard next-token prediction objective on the mixed data and use the resulting policy $\pi_{\mathrm{SFT}}$ to initialize HCD-GRPO.

\subsection{Hierarchical Coverage-Discounted GRPO}
\label{sec:binding-policy-optimization}

Starting from $\pi_{\mathrm{SFT}}$, the policy samples multiple candidate captions for each video--reference input. We score these candidates, normalize their rewards within each group, and update the policy using the standard clipped GRPO objective. During this stage, we freeze the visual encoder and vision--language merger and update only the language model.

HCD-GRPO uses two reward branches. The caption branch evaluates whether the output describes the video correctly and completely. The reference branch evaluates whether the output selects, binds, and groups the reference images correctly. Both branches follow a coverage-discounted form: useful coverage provides the positive signal, while observable errors discount that signal. We combine the two branch rewards using fixed weights and cap the resulting reward when the output contains malformed or nonexistent image tags.

\subsubsection{Factual Caption Reward}

We first remove all image tags from the generated caption to obtain $\bar{y}=\operatorname{StripTags}(y)$. A keypoint bank measures its coverage of the subjects, appearance, actions, background, camera motion, and visual style. For each keypoint, a structured LLM judge assigns $s_k\in\{0,0.5,1\}$ for missing or incorrect, partial, or correct coverage. A separate video-QA bank checks factual errors and assigns $e_q\in\{0,0.5,1\}$ for supported or unanswered, partially incorrect, or incorrect content. We define the factual caption reward as
\begin{equation}
R_{\mathrm{cap}}(x,y)
=
\left(\frac{1}{K}\sum_{k=1}^{K}s_k\right)
\left[
1-\lambda_{\mathrm{qa}}
\left(\frac{1}{Q}\sum_{q=1}^{Q}e_q\right)
\right]_+,
\label{eq:caption-reward}
\end{equation}
where $[z]_+=\max(0,z)$. The first term rewards factual coverage, while the second discounts incorrect content. Unanswered questions receive no additional error penalty because omissions are already reflected in keypoint coverage.

\subsubsection{Multi-Reference Grounding Reward}

\noindent\textbf{\textit{Correctly Bound Reference Coverage Reward.}}
Let $\mathcal{T}^{+}(x)$ and $\mathcal{T}^{-}(x)$ denote the valid and distractor reference tags, respectively, and let $\mathcal{T}_{y}$ denote the distinct tags used in the candidate caption. We define $C_{\mathrm{ref}}$ as the fraction of valid references that are both emitted and bound to semantically corresponding phrases. A reference contributes to this coverage only when its item-level binding decision is correct. This design prevents an incorrectly bound tag from increasing the positive reward, while still discouraging the model from avoiding difficult binding decisions by emitting few or no image tags.

\noindent\textbf{\textit{Reference Binding Accuracy Reward.}}
Each reference image $I_i$ is paired with an annotated image description $d_i$. We parse the candidate caption into local phrase--tag pairs $(p_j,t_j)$, where $p_j$ is the phrase immediately preceding the emitted tag $t_j=\texttt{<Image\_$i_j$>}$. For each pair, an LLM judge assigns a binary score $a_j$, where $a_j=1$ indicates that $p_j$ and $d_{i_j}$ describe the same visual content and $a_j=0$ indicates an incorrect binding. The binding accuracy is
\begin{equation}
A_{\mathrm{bind}}(x,y)
=
\frac{1}{|\mathcal{B}_{y}|}
\sum_{(p_j,t_j)\in\mathcal{B}_{y}}a_j,
\label{eq:binding-accuracy}
\end{equation}
where $\mathcal{B}_{y}$ contains all valid phrase--tag pairs in the candidate caption. We set $A_{\mathrm{bind}}=0$ when the caption contains no valid image tag.

\noindent\textbf{\textit{Distractor-Aware Evidence Suppression (DAES).}}
DAES checks whether the caption uses a reference whose content does not appear in the video. We set $E_{\mathrm{DAES}}=1$ when the candidate uses any distractor tag and $E_{\mathrm{DAES}}=0$ otherwise. This term penalizes the model for treating an irrelevant reference as visual evidence.

\noindent\textbf{\textit{Cross-Reference Semantic Coherence (CRSC).}}
The annotations specify which reference images depict the same entity across different viewpoints, poses, or appearances. The entities may be people, objects, or backgrounds. Let $\mathcal{E}_{\mathrm{multi}}(x)$ denote the annotated entities associated with multiple reference images, and let $\mathcal{T}^{*}_{e}$ denote the ground-truth tag set for entity $e$. For each annotated entity, an LLM judge locates the corresponding entity phrase in the candidate caption and extracts the image tags placed immediately after that phrase. We denote the extracted tag set by $\widehat{\mathcal{T}}_{e}(y)$. Tags attached to separate entity phrases are not merged. We compare the extracted and annotated tag sets to compute the entity-level consistency accuracy:
\begin{equation}
A_{\mathrm{CRSC}}(x,y)
=
\frac{1}{|\mathcal{E}_{\mathrm{multi}}(x)|}
\sum_{e\in\mathcal{E}_{\mathrm{multi}}(x)}
\frac{
|\widehat{\mathcal{T}}_{e}(y)
\cap
\mathcal{T}^{*}_{e}|
}{
|\widehat{\mathcal{T}}_{e}(y)
\cup
\mathcal{T}^{*}_{e}|
}.
\label{eq:crsc-accuracy}
\end{equation}
The score reaches its maximum only when all references belonging to the same entity are grouped after the corresponding entity phrase. It decreases when the model omits an expected tag, adds an unrelated tag, or splits different views of the same entity across separate phrases.

\begin{figure*}[!t]
  \centering
  \includegraphics[width=\textwidth]{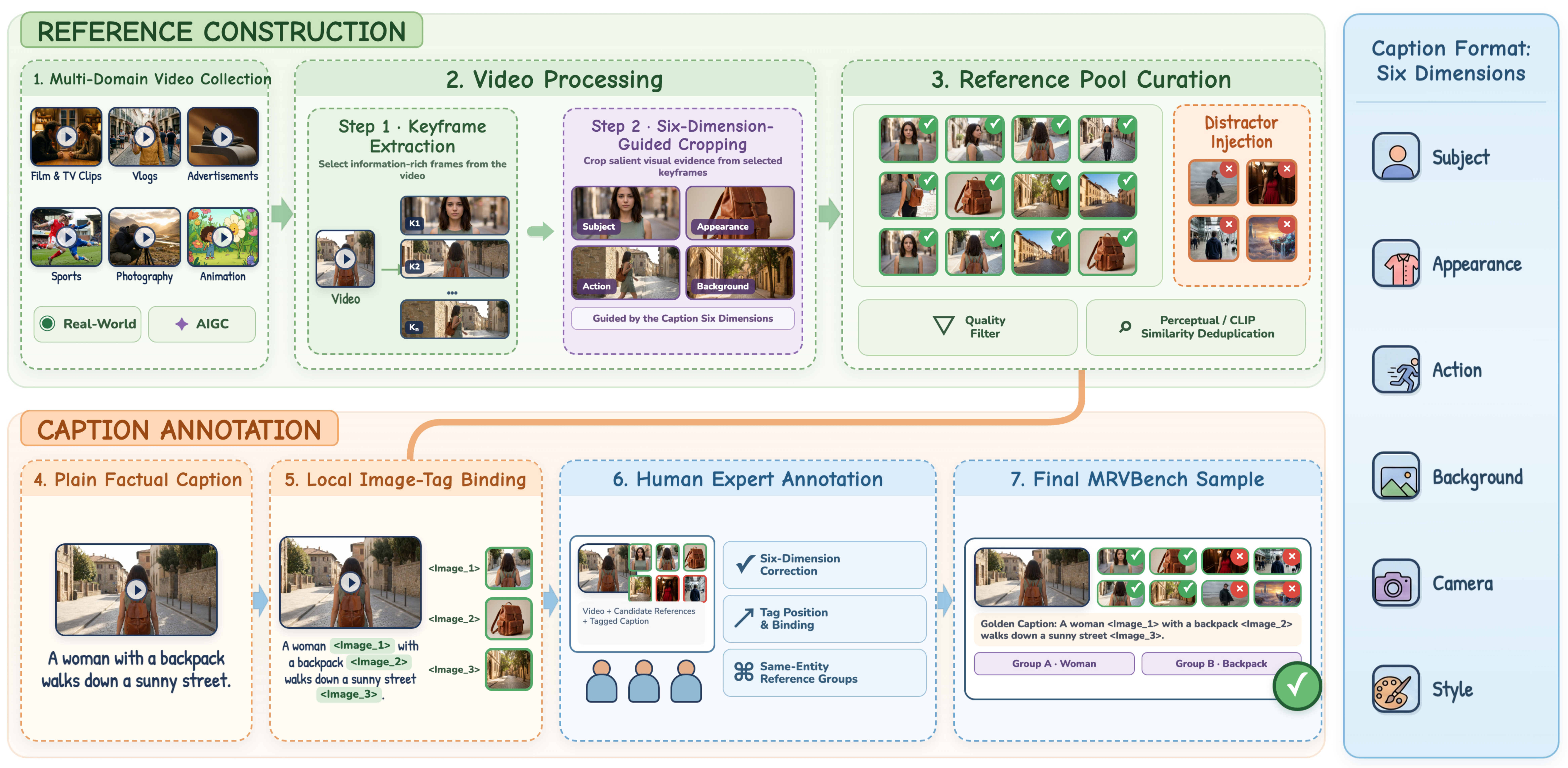}
  \caption{MRVBench data construction pipeline. The pipeline includes multi-domain video collection, keyframe extraction, six-dimension-guided cropping, reference pool curation with distractor injection, local image-tag binding, and human verification.}
  \label{fig:supp-mrvbench-construction}
\end{figure*}

Combining these components, the multi-reference grounding reward is
\begin{equation}
\begin{aligned}
R_{\mathrm{ref}}(x,y)
&=C_{\mathrm{ref}}(x,y)\Big[
1-\lambda_b\left(1-A_{\mathrm{bind}}\right) \\
&\quad-m_d\lambda_dE_{\mathrm{DAES}}
-m_c\lambda_c\left(1-A_{\mathrm{CRSC}}\right)
\Big]_+,
\end{aligned}
\label{eq:reference-reward}
\end{equation}
where $m_d$ activates DAES only for samples containing distractors, and $m_c$ activates CRSC only for samples containing an entity with multiple references.

\section{MRVBench}
\label{sec:mrvbench}

We introduce \textbf{MRVBench} to jointly evaluate video factuality, reference selection, and phrase-level reference binding, and construct a large-scale training corpus and a compact human-verified benchmark through MLLM-assisted annotation. The training corpus contains \textbf{20,000 videos} and \textbf{171,354 reference images} collected from authorized sources under appropriate usage rights. The test set comprises \textbf{462 videos}---\textbf{185} AIGC videos (40\%) and \textbf{277} real-world videos (60\%)---together with \textbf{3,831 reference images}. All 462 test samples are fully held out from the training corpus and are used only for evaluation. Difficulty increases with the numbers of references and visual units, mapping complexity, and distractors: a video can have up to \textbf{22 reference images}, and each sample has at least 3 video key-point annotations and 4 video logical questions. Approximately \textbf{60\%} of the samples contain complex mappings in which multiple references correspond to the same subject or visual unit, while approximately \textbf{40\%} (about 185 samples) contain injected distractors. Data-source, construction, and evaluation details are provided in Appendix~\ref{sec:supp-mrvbench}.

\subsection{Evaluation Metrics}
\label{sec:evaluation-protocol}

We evaluate multi-reference video captioning from the following dimensions. \textbf{KP-Cov} measures how completely a caption covers human-verified video facts. \textbf{VQA} measures whether the caption supports correct answers to video questions, while \textbf{VQA-Cov} measures the proportion of questions covered by the caption. \textbf{Ref-Tag-P/R} measure the precision and recall of the selected reference images. \textbf{Ref-Bind} measures whether emitted image tags are attached to the correct local phrases, while \textbf{Eff-Bind} jointly measures reference selection and phrase-level binding. \textbf{Dist-Rej} measures the proportion of irrelevant references excluded from a caption, and \textbf{FalseRef-Any} measures the proportion of samples containing at least one incorrectly used distractor. \textbf{Subj-R/F1} evaluate whether multiple references depicting the same subject are consistently grouped with the corresponding entity. We use \textbf{Gemini-3.1-Pro} as the MLLM judge. 
The details of the evaluation are in Appendix~\ref{sec:supp-mrvbench-metrics}.

\begin{table*}[t]
  \centering
  \caption{\textbf{Main results on MRVBench.}
  The best result in each column is highlighted in \colorbox{mrvgreen}{green}.
  Gray rows show proprietary models for reference and are excluded from best-value highlighting.}
  \label{tab:mrvbench-main-results}

  \begingroup
  \small
  \setlength{\tabcolsep}{0.600pt}
  \renewcommand{\arraystretch}{1.1}

  \resizebox{\textwidth}{!}{%
  \begin{tabular}{lcccccccccccc}
    \toprule
    \multirow{3}{*}{\textbf{Model}}
    & \multicolumn{3}{c}{\textbf{Caption and Video Content}}
    & \multicolumn{8}{c}{\textbf{Multi-Reference Grounding}}
    & \multirow{3}{*}{\textbf{MRVScore}$\uparrow$} \\
    \cmidrule(lr){2-4}
    \cmidrule(lr){5-12}
    & \multicolumn{1}{c}{\shortstack{\textbf{Key-Point}\\\textbf{Coverage}}}
    & \multicolumn{2}{c}{\textbf{Video Factuality}}
    & \multicolumn{2}{c}{\textbf{Reference Selection}}
    & \multicolumn{2}{c}{\textbf{Reference Binding}}
    & \multicolumn{2}{c}{\textbf{Distractor Robustness}}
    & \multicolumn{2}{c}{\textbf{Subject Consistency}}
    & \\
    \cmidrule(lr){2-2}
    \cmidrule(lr){3-4}
    \cmidrule(lr){5-6}
    \cmidrule(lr){7-8}
    \cmidrule(lr){9-10}
    \cmidrule(lr){11-12}
    & \textbf{KP-Cov}$\uparrow$
    & \textbf{VQA}$\uparrow$
    & \textbf{VQA-Cov}$\uparrow$
    & \textbf{Ref-Tag-P}$\uparrow$
    & \textbf{Ref-Tag-R}$\uparrow$
    & \textbf{Ref-Bind}$\uparrow$
    & \textbf{Eff-Bind}$\uparrow$
    & \textbf{Dist-Rej}$\uparrow$
    & \textbf{FalseRef-Any}$\downarrow$
    & \textbf{Subj-R}$\uparrow$
    & \textbf{Subj-F1}$\uparrow$
    & \\
    \midrule

    \rowcolor{mrvblue}
    \multicolumn{13}{c}{\textbf{Proprietary Models}} \\
    {\hypersetup{citecolor=gray}\textcolor{gray}{Gemini-3.1-Pro}\hypersetup{citecolor=cvprblue}}
    & \textcolor{gray}{0.878} & \textcolor{gray}{0.751} & \textcolor{gray}{0.879} & \textcolor{gray}{0.992} & \textcolor{gray}{0.938} & \textcolor{gray}{0.976} & \textcolor{gray}{0.915} & \textcolor{gray}{0.978} & \textcolor{gray}{0.100} & \textcolor{gray}{0.799} & \textcolor{gray}{0.844} & \textcolor{gray}{0.897} \\
    {\hypersetup{citecolor=gray}\textcolor{gray}{GPT-5.4}\hypersetup{citecolor=cvprblue}}
    & \textcolor{gray}{0.927} & \textcolor{gray}{0.800} & \textcolor{gray}{0.923} & \textcolor{gray}{0.998} & \textcolor{gray}{0.807} & \textcolor{gray}{0.986} & \textcolor{gray}{0.795} & \textcolor{gray}{0.993} & \textcolor{gray}{0.0359} & \textcolor{gray}{0.609} & \textcolor{gray}{0.718} & \textcolor{gray}{0.870} \\

    \midrule
    \rowcolor{mrvpink}
    \multicolumn{13}{c}{\textbf{Open-source General-purpose Models}} \\
    ARC-Hunyuan-Video-7B~\cite{ge2025archunyuan}
    & 0.528 & 0.377 & 0.526 & 0.0779 & 0.0376 & 0.920 & 0.0346 & 0.980 & 0.122 & 0.0217 & 0.0420 & 0.325 \\
    InternVL3.5-38B~\cite{wang2025internvl3}
    & 0.729 & 0.562 & 0.721 & 0.957 & 0.760 & 0.902 & 0.686 & 0.905 & 0.315 & 0.514 & 0.645 & 0.746 \\
    InternVL3.5-8B~\cite{wang2025internvl3}
    & 0.735 & 0.553 & 0.719 & 0.569 & 0.316 & 0.857 & 0.271 & 0.967 & 0.160 & 0.214 & 0.329 & 0.557 \\
    Keye-VL-1.5-8B~\cite{yang2025keyevl15}
    & 0.766 & 0.583 & 0.740 & 0.423 & 0.237 & 0.723 & 0.171 & 0.950 & 0.195 & 0.108 & 0.186 & 0.507 \\
    MiMo-VL-7B-RL~\cite{xiaomi2025mimo}
    & 0.740 & 0.572 & 0.731 & 0.785 & 0.475 & 0.889 & 0.422 & 0.983 & 0.156 & 0.233 & 0.357 & 0.633 \\
    Qwen3-VL-32B-Instruct~\cite{bai2025qwen3vl}
    & 0.849 & \cellcolor{mrvgreen}\textbf{0.701} & 0.830 & 0.979 & 0.867 & 0.891 & 0.773 & 0.941 & 0.205 & 0.599 & 0.695 & 0.829 \\
    Qwen3-VL-32B-Thinking~\cite{bai2025qwen3vl}
    & 0.599 & 0.505 & 0.654 & 0.751 & 0.711 & 0.340 & 0.242 & 0.755 & 0.670 & 0.364 & 0.517 & 0.592 \\
    Qwen3-VL-8B-Instruct~\cite{bai2025qwen3vl}
    & 0.825 & 0.670 & 0.814 & 0.940 & 0.711 & 0.842 & 0.599 & 0.948 & 0.210 & 0.471 & 0.599 & 0.763 \\
    Qwen3-VL-8B-Thinking~\cite{bai2025qwen3vl}
    & 0.576 & 0.466 & 0.629 & 0.846 & 0.904 & 0.198 & 0.179 & 0.633 & 0.900 & 0.634 & 0.733 & 0.621 \\
    Qwen3.6-27B~\cite{team2026qwen327b}
    & 0.835 & 0.685 & 0.831 & 0.979 & 0.790 & 0.966 & 0.763 & 0.974 & 0.100 & 0.503 & 0.656 & 0.814 \\
    Qwen3.6-35B-A3B~\cite{team2026qwen3}
    & 0.836 & 0.684 & 0.827 & 0.867 & 0.555 & 0.951 & 0.528 & 0.957 & 0.155 & 0.333 & 0.491 & 0.718 \\
    \midrule
    \rowcolor{mrvyellow}
    \multicolumn{13}{c}{\textbf{Our Model}} \\
    \textbf{RefCaptioner}
    & \cellcolor{mrvgreen}\textbf{0.882}
    & 0.686
    & \cellcolor{mrvgreen}\textbf{0.837}
    & \cellcolor{mrvgreen}\textbf{0.994}
    & \cellcolor{mrvgreen}\textbf{0.943}
    & \cellcolor{mrvgreen}\textbf{0.967}
    & \cellcolor{mrvgreen}\textbf{0.912}
    & \cellcolor{mrvgreen}\textbf{0.985}
    & \cellcolor{mrvgreen}\textbf{0.100}
    & \cellcolor{mrvgreen}\textbf{0.817}
    & \cellcolor{mrvgreen}\textbf{0.869}
    & \cellcolor{mrvgreen}\textbf{0.888} \\
    \bottomrule
  \end{tabular}}
  \endgroup
\end{table*}

\section{Experiments}
\label{sec:experiments}

\subsection{Experimental settings}
  \label{sec:experimental-setup}

  RefCaptioner was trained in two stages: supervised fine-tuning (SFT) followed by GRPO-based
  post-training. The best-performing SFT checkpoint on the validation set was used to initialize
  GRPO. Detailed training configurations and optimization hyperparameters are provided in the
  Appendix~\ref{sec:supp-training}. Our experiments utilized 32 NVIDIA H800 80GB GPUs

\textbf{Implementation details} We initialized RefCaptioner from \textbf{Qwen3-VL-8B-Instruct}. During SFT, LoRA adapters were
applied to eligible linear layers while the backbone remained frozen. During GRPO, we optimized
the full LLM while keeping the Vision encoder and Vision-Language Fusion modules
frozen. The details of HCD-GRPO are provided in Appendix~\ref{sec:supp-grpo}--\ref{sec:supp-reward-configuration}.

\textbf{Baselines} We compared RefCaptioner with ARC-Hunyuan-Video-7B~\cite{ge2025archunyuan},
InternVL3.5-38B~\cite{wang2025internvl3},
Keye-VL-1.5-8B~\cite{yang2025keyevl15},
LLaVA-OneVision-2-8B-Instruct~\cite{an2026llavaonevision2},
MiMo-VL-7B-RL~\cite{xiaomi2025mimo}, multiple variants from the
Qwen3-VL~\cite{bai2025qwen3vl} and
Qwen3.6~\cite{team2026qwen327b,team2026qwen3}
series, VideoLLaMA3-7B~\cite{zhang2025videollama3}, and
LLaVA-Video-7B~\cite{zhang2024llavavideo}. We further included
Gemini-3.1-Pro and
GPT-5.4 as strong proprietary baselines. The general video-captioning evaluation settings are provided in Appendix~\ref{sec:supp-general-evaluation}.

\subsection{Main Results}
\label{sec:mrvbench-main-results}

\textbf{Standard video-caption evaluation.}
As shown in Table~\ref{tab:mrvbench-main-results}, RefCaptioner achieves the best overall performance among the evaluated open-source models, with an MRVScore close to Gemini-3.1-Pro and higher than GPT-5.4. On standard video-caption evaluation, it obtains the highest KP-Cov and VQA-Cov among open-source models, while its KP-Cov is also comparable to the proprietary baselines. RefCaptioner further improves VQA over its Qwen3-VL-8B-Instruct base model, although the larger Qwen3-VL-32B-Instruct performs slightly better on this metric. These results show that RefCaptioner maintains factual video understanding while providing more complete descriptions.

\textbf{Multi-reference grounding evaluation.}
RefCaptioner shows a clearer advantage on multi-reference grounding. It achieves the best or tied-best result on every reported grounding metric among open-source models and remains competitive with the proprietary systems. In particular, it surpasses both Gemini-3.1-Pro and GPT-5.4 on Ref-Tag-R, Subj-R, and Subj-F1, and exceeds at least one of them on Ref-Tag-P, Eff-Bind, and Dist-Rej. Its strong Ref-Tag-P and Ref-Tag-R show that it selects valid references without omitting them; Ref-Bind and Eff-Bind confirm that the selected images are attached to the correct phrases; and the distractor and subject-consistency metrics show that it rejects irrelevant images while grouping multiple views of the same subject. RefCaptioner therefore provides a stronger balance between reference coverage, local binding, distractor rejection, and subject consistency.

\subsection{Comparison with Two-Stage Caption Refinement}
\label{sec:captionrefine-comparison}

\begin{table}[t]
  \centering
  \caption{\textbf{Comparison with two-stage caption refinement on MRVBench.}
  Each baseline first generates a video-only draft and then revises it using the video and ordered reference images. RefCaptioner directly generates a grounded caption in one pass.}
  \label{tab:captionrefine-comparison}
  \begingroup
  \small
  \setlength{\tabcolsep}{1.0pt}
  \renewcommand{\arraystretch}{1.18}
  \begin{tabular*}{\columnwidth}{@{\extracolsep{\fill}}lccccc@{}}
    \toprule
    \textbf{Model}
    & \shortstack{\textbf{KP}\\\textbf{Cov}$\uparrow$}
    & \shortstack{\textbf{Ref-Tag}\\\textbf{R}$\uparrow$}
    & \shortstack{\textbf{Ref}\\\textbf{Bind}$\uparrow$}
    & \shortstack{\textbf{Subj}\\\textbf{R}$\uparrow$}
    & \shortstack{\textbf{Subj}\\\textbf{F1}$\uparrow$} \\
    \midrule
    \multicolumn{6}{l}{\textit{Two-Stage CaptionRefine}} \\
    Qwen3-VL-8B
    & 0.868 & 0.508 & 0.366 & 0.337 & 0.366 \\
    Qwen3-VL-32B
    & \textbf{0.911} & 0.869 & 0.840 & 0.638 & 0.707 \\
    InternVL3.5-8B
    & 0.728 & 0.486 & 0.229 & 0.369 & 0.359 \\
    InternVL3.5-38B
    & 0.732 & 0.655 & 0.805 & 0.365 & 0.490 \\
    \midrule
    \textbf{RefCaptioner (Ours)}
    & 0.882
    & \textbf{0.943}
    & \textbf{0.967}
    & \textbf{0.817}
    & \textbf{0.869} \\
    \bottomrule
  \end{tabular*}
  \endgroup
\end{table}

To examine whether reference grounding can be recovered through post-hoc rewriting rather than learned during post-training, we compare RefCaptioner with a two-stage CaptionRefine pipeline. Each baseline first generates a standard caption from the video alone. The same model then receives the video, the ordered reference images, and its initial caption, and revises the caption by inserting the corresponding \texttt{<Image\_n>} tags. Only the revised captions are evaluated, using the same 462 MRVBench samples for all methods. As shown in Table~\ref{tab:captionrefine-comparison}, Qwen3-VL-32B-Instruct with CaptionRefine obtains the highest KP-Cov, indicating that an additional pass can preserve broad coverage of the video content. RefCaptioner, however, performs best on Ref-Tag-R, Ref-Bind, Subj-R, and Subj-F1, with clear margins over every two-stage baseline. This contrast shows that post-hoc revision can add reference tags to a descriptive caption, but does not reliably recover phrase-level reference correspondences or group multiple images of the same subject. Reference selection and binding are therefore better learned together with caption generation than treated as a later correction step.

\subsection{General Video-Captioning Performance}
\label{sec:general-captioning-performance}

To determine whether reference-oriented post-training preserves general video-captioning ability, we compare RefCaptioner with open-source models on VDC and VCapsBench. As shown in Table~\ref{tab:vdc-results}, RefCaptioner achieves the highest accuracy across all five VDC dimensions. The largest gains over its Qwen3-VL-8B base model occur in background, main-object, and detailed description, where accuracy improves by 5.66--6.08 points. This result shows that the model does not specialize only in image-tag insertion, but also captures ordinary video content more completely. Table~\ref{tab:vcapsbench-results} shows a similar trend: RefCaptioner obtains the highest AR and CR, improving them over its base model by 1.30 and 1.70 percentage points, respectively, while maintaining a comparable IR. MiniCPM-V4.5 achieves a lower IR, but its AR and CR remain substantially below those of RefCaptioner. Taken together, the two benchmarks show that multi-reference post-training improves detailed caption generation while maintaining a strong balance between content coverage and caption consistency.

\begin{table}[t]
  \centering
  \caption{\textbf{Results on VDC.}
  We report accuracy across the five evaluation dimensions; higher is better. LLaVA-OV denotes LLaVA-OneVision.}
  \label{tab:vdc-results}
  \begingroup
  \small
  \setlength{\tabcolsep}{0.5pt}
  \renewcommand{\arraystretch}{1.2}
  \begin{tabular*}{\columnwidth}{@{\extracolsep{\fill}}lccccc@{}}
    \toprule
    \textbf{Model}
    & \textbf{Cam.}$\uparrow$
    & \textbf{Short}$\uparrow$
    & \textbf{Back.}$\uparrow$
    & \textbf{Obj.}$\uparrow$
    & \textbf{Detail.}$\uparrow$ \\
    \midrule
    \multicolumn{6}{l}{\textit{Open-Source Models}} \\
    Qwen3-VL-8B-Inst.~\cite{bai2025qwen3vl}
    & 57.44 & 46.91 & 67.57 & 67.74 & 64.03 \\
    InternVL3.5-8B~\cite{wang2025internvl3}
    & 27.96 & 34.83 & 40.73 & 39.95 & 34.22 \\
    Keye-VL-1.5-8B~\cite{yang2025keyevl15}
    & 51.47 & 42.53 & 59.65 & 60.12 & 56.58 \\
    LLaVA-OV1.5-8B
    & 38.66 & 40.24 & 49.85 & 50.06 & 45.15 \\
    LLaVA-OV2-8B~\cite{an2026llavaonevision2}
    & 41.23 & 38.40 & 49.03 & 49.04 & 43.25 \\
    MiniCPM-V4.5-8B
    & 27.00 & 28.98 & 32.03 & 32.52 & 25.45 \\
    MiMo-VL-7B-RL~\cite{xiaomi2025mimo}
    & 50.16 & 41.31 & 55.66 & 55.29 & 52.72 \\
    \midrule
    \textbf{RefCaptioner (Ours)}
    & \textbf{59.54}
    & \textbf{50.87}
    & \textbf{73.23}
    & \textbf{73.50}
    & \textbf{70.11} \\
    \bottomrule
  \end{tabular*}
  \endgroup
\end{table}

\begin{center}
  \centering
  \captionof{table}{\textbf{Results on VCapsBench.}
  AR and CR are higher-is-better; IR is lower-is-better.}
  \label{tab:vcapsbench-results}
  \begingroup
  \small
  \setlength{\tabcolsep}{1.2pt}
  \renewcommand{\arraystretch}{1.2}
  \begin{tabular*}{\columnwidth}{@{\extracolsep{\fill}}lcccc@{}}
    \toprule
    \textbf{Model} & \textbf{Size}
    & \textbf{AR}$\uparrow$
    & \textbf{IR}$\downarrow$
    & \textbf{CR}$\uparrow$ \\
    \midrule
    \multicolumn{5}{l}{\textit{Open-Source Models}} \\
    Qwen3-VL-8B-Inst.~\cite{bai2025qwen3vl}
    & 8B & 64.83 & 13.19 & 74.69 \\
    InternVL3.5-8B~\cite{wang2025internvl3}
    & 8B & 54.22 & 14.19 & 63.18 \\
    Keye-VL-1.5-8B~\cite{yang2025keyevl15}
    & 8B & 55.58 & 13.13 & 63.98 \\
    LLaVA-OV1.5-8B
    & 8B & 52.93 & 17.01 & 63.78 \\
    LLaVA-OV2-8B~\cite{an2026llavaonevision2}
    & 8B & 54.04 & 12.54 & 61.78 \\
    MiniCPM-V4.5
    & 8B & 60.39 & \textbf{12.46} & 68.99 \\
    MiMo-VL-7B-RL~\cite{xiaomi2025mimo}
    & 7B & 51.32 & 13.19 & 59.11 \\
    \midrule
    \textbf{RefCaptioner (Ours)}
    & \textbf{8B}
    & \textbf{66.13}
    & 13.42
    & \textbf{76.39} \\
    \bottomrule
  \end{tabular*}
  \endgroup
\end{center}

\begin{center}
  \centering
  \includegraphics[width=0.90\columnwidth]{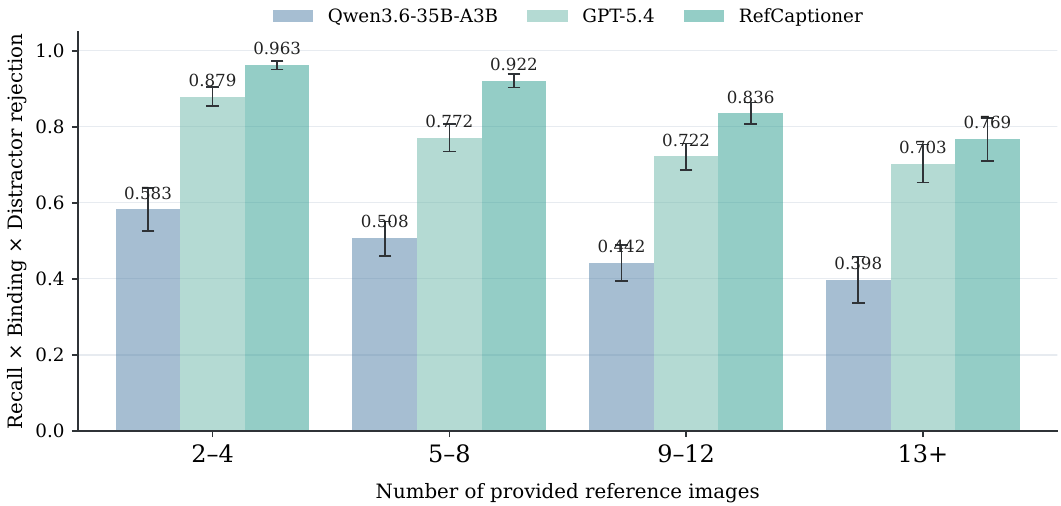}
  \captionof{figure}{Reference-grounding robustness under increasing numbers of reference images. The robustness score is the product of reference recall, binding accuracy, and distractor rejection.}
  \label{fig:grounding-robustness-by-reference-count}
\end{center}

\begin{figure*}[!htbp]
  \centering
  \includegraphics[width=\textwidth]{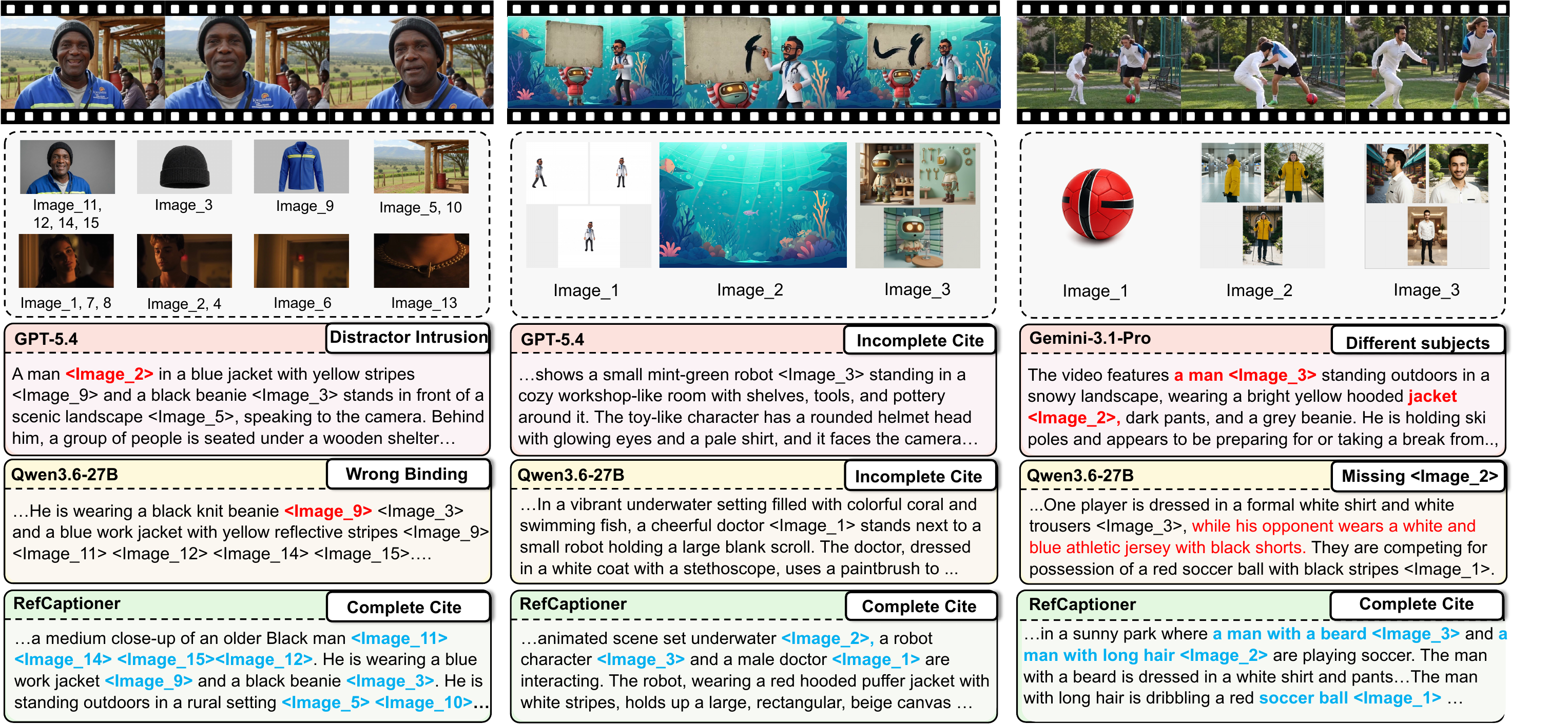}
  \caption{\textbf{Qualitative comparison on MRVBench.}
  Baseline captions exhibit distractor intrusion, incorrect or incomplete reference binding, and inconsistent subject grouping. RefCaptioner selects the supported references and places their tags after the corresponding visual phrases.}
  \label{fig:qualitative-comparison}
\end{figure*}

\subsection{Robustness to Increasing Reference Images}
\label{sec:reference-complexity-analysis}

We group MRVBench samples by reference count: 2--4, 5--8, 9--12, and 13+. To jointly measure correct reference selection, phrase-level binding, and distractor exclusion, we compute a reference-grounding robustness score as the product of reference recall, binding accuracy, and distractor rejection within each group; this multiplicative form penalizes models that are weak in any one component. As shown in Figure~\ref{fig:grounding-robustness-by-reference-count}, RefCaptioner achieves the highest robustness score across all reference-count groups. On the most challenging 13+ group, it obtains a score of 0.769, outperforming GPT-5.4 (0.703) and Qwen3.6-35B-A3B (0.398), demonstrating stronger robustness to complex reference sets.

\subsection{Qualitative Analysis}
\label{sec:qualitative-analysis}

Figure~\ref{fig:qualitative-comparison} compares three representative examples. In the left example, GPT-5.4 uses the distractor \texttt{<Image\_2>}, while Qwen3.6-27B swaps the jacket, beanie, and subject tags. RefCaptioner instead groups the four views of the same man and correctly binds the clothing and background references. In the middle example, GPT-5.4 hallucinates a workshop and cites only the robot, while Qwen3.6-27B cites the doctor but misses the robot and underwater scene. RefCaptioner binds all three elements. In the right example, Gemini-3.1-Pro treats the second man as an attribute of the first, and Qwen3.6-27B omits his reference. RefCaptioner separates the two players and correctly binds the ball. These cases demonstrate improved reference recall, local binding, and subject consistency. An additional qualitative example is provided in Appendix~\ref{sec:supp-mrvbench-qualitative}.

\begin{figure}[!htbp]
  \centering
  \includegraphics[width=0.90\columnwidth]{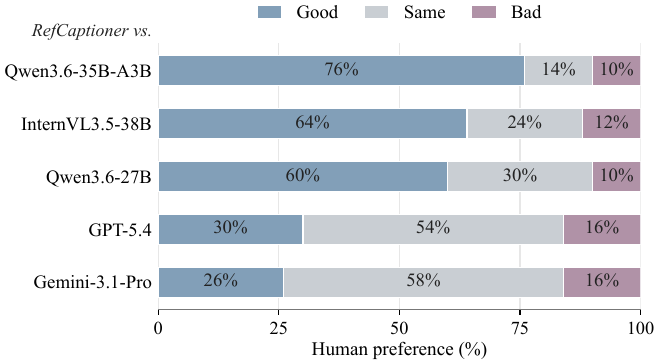}
  \caption{Pairwise Good/Same/Bad (GSB) evaluation comparing RefCaptioner with different baselines for caption-conditioned video reconstruction.}
  \label{fig:reconstruction-gsb}
\end{figure}

\subsection{Caption-Conditioned Video Reconstruction}
\label{sec:caption-conditioned-reconstruction}

To assess downstream utility, we reconstruct videos from each generated caption and its cited reference images under identical generation settings. Human evaluators consistently prefer RefCaptioner-based reconstructions. As shown in Figure~\ref{fig:reconstruction-gsb}, RefCaptioner receives more Good than Bad judgments against every baseline, with particularly pronounced advantages over the open-source models. These results indicate that RefCaptioner better preserves source-video information through its generated descriptions and selected references. Additional video-reconstruction case studies are provided in Appendix~\ref{sec:supp-reconstruction-comparisons}.

\subsection{Ablation Study}
\label{sec:ablation-study}

Table~\ref{tab:ablation-study} studies the two training stages and three HCD-GRPO reward components. SFT substantially improves subject recall over the base model but reduces VQA, indicating that supervised multi-reference learning alone favors reference handling at the expense of factual description. HCD-GRPO resolves this imbalance, with the full model achieving the best VQA, Dist-Rej, and Subj-R among all variants. Removing the factual-caption reward causes the largest VQA reduction, removing DAES most strongly affects distractor rejection, and removing CRSC produces a smaller but consistent decrease in subject recall. The three rewards therefore provide complementary supervision for factual captioning, distractor rejection, and cross-reference consistency.

\begin{center}
  \centering
  \begingroup
  \small
  \setlength{\tabcolsep}{1.4pt}
  \renewcommand{\arraystretch}{1.12}
  \begin{tabular}{lccc|ccc}
    \toprule
    \multirow{2}{*}{\textbf{Variant}}
    & \multicolumn{3}{c|}{\textbf{Reward}}
    & \multicolumn{3}{c}{\textbf{Metrics}} \\
    \cmidrule(lr){2-4}
    \cmidrule(lr){5-7}
    & \textbf{Fact.}
    & \textbf{DAES}
    & \textbf{CRSC}
    & \textbf{VQA}$\uparrow$
    & \textbf{Dist-Rej}$\uparrow$
    & \textbf{Subj-R}$\uparrow$ \\
    \midrule
    Base
    & -- & -- & --
    & 0.670 & 0.948 & 0.471 \\
    SFT
    & -- & -- & --
    & 0.642 & 0.933 & 0.702 \\
    \midrule
    \multicolumn{7}{c}{\textit{HCD-GRPO Variants}} \\
    \cmidrule(lr){1-7}
    w/o Factual
    & -- & \checkmark & \checkmark
    & 0.642 & 0.934 & 0.770 \\
    w/o DAES
    & \checkmark & -- & \checkmark
    & 0.663 & 0.914 & 0.740 \\
    w/o CRSC
    & \checkmark & \checkmark & --
    & 0.667 & 0.931 & 0.778 \\
    \textbf{Full}
    & \checkmark & \checkmark & \checkmark
    & \textbf{0.686} & \textbf{0.985} & \textbf{0.817} \\
    \bottomrule
  \end{tabular}
  \endgroup
  \captionof{table}{Ablation of training stages and reward components. The three HCD-GRPO ablations remove the factual-caption reward, DAES, or CRSC, respectively.}
  \label{tab:ablation-study}
\end{center}

\section{Conclusion}

We introduce RefCaptioner for multi-reference image-grounded video captioning, combining mixed-data SFT with HCD-GRPO to optimize factual captioning and reliable reference grounding. 
We also present MRVBench for evaluating video factuality and multi-reference grounding on real-world and AI-generated videos. 
RefCaptioner achieves the best overall performance among open-source models while remaining strong on general captioning benchmarks. 
Human evaluation validates its caption quality and phrase-level binding accuracy, and controlled reconstruction provides initial evidence of downstream utility. 
Overall, explicit multi-reference grounding can be incorporated into video captioning without compromising general description quality.

{
    \small
    \bibliographystyle{ieeenat_fullname}
    \bibliography{main}
}

\clearpage
\maketitlesupplementary

\makeatletter
\setlength{\@dblfptop}{0pt}
\makeatother

\appendix
\setcounter{figure}{0}
\setcounter{table}{0}
\setcounter{equation}{0}
\renewcommand{\thefigure}{S\arabic{figure}}
\renewcommand{\thetable}{S\arabic{table}}
\renewcommand{\theequation}{S\arabic{equation}}
\renewcommand{\theHfigure}{S.\arabic{figure}}
\renewcommand{\theHtable}{S.\arabic{table}}
\renewcommand{\theHequation}{S.\arabic{equation}}

\section{MRVBench Construction and Evaluation}
\label{sec:supp-mrvbench}

\subsection{Benchmark Overview}

MRVBench contains 462 test samples, comprising 185 AIGC videos and 277 real videos, with 3,831 candidate reference images in total. Each sample consists of one video, multiple tagged candidate reference images, a reference-grounded golden caption, per-reference keep/delete annotations, same-entity reference-group annotations, a frozen key-point bank, and a frozen video-QA bank. Candidate references include both positive references and distractors. A positive reference contains a useful visual element that can be grounded in the video, whereas a distractor does not provide a valid grounding target.

\subsection{Data Sources and Release Scope}

The training corpus was lawfully collected from authorized sources under appropriate usage rights. Because redistribution rights differ across these sources, the training corpus will not be publicly released. We will release the MRVBench test set and its evaluation annotations as a benchmark for research use, subject to the applicable data licenses. The release will include only data and annotations that can be redistributed under those licenses.

\subsection{Annotation Structure}

Each positive reference is assigned to a concrete visual unit in the video, such as a subject, object, appearance attribute, or background element. References share a group only when they depict the same entity or visual unit across viewpoints, poses, or appearances; semantically related but distinct entities are not merged. Each reference additionally includes a grounded phrase and concise visual evidence supporting its keep/delete decision.

The frozen key-point bank is derived from tag-free golden captions and organizes independently checkable facts into six dimensions: subject, appearance, action, background, camera, and style. The 462 samples contain 5,846 key points. The frozen QA bank contains 2,172 short-answer questions covering entity actions, entity--background relations, multi-entity interactions, camera behavior, appearance, and scene style. Each question includes a reference answer and its supporting evidence dimension. These annotation banks are generated offline and remain fixed for all evaluated models.

\subsection{Metric Definitions}
\label{sec:supp-mrvbench-metrics}

Let $C_i$ be the set of unique image tags used by a candidate caption, $G_i$ the set of positive tags, and $N_i$ the set of distractor tags for sample $i$.

\noindent\textbf{Key-Point Coverage (KP-Cov).}
Every key point is labeled as correct, partial, or missing. KP-Cov gives coverage credit to correct and partial matches and pools counts over the test set:
\begin{equation}
\mathrm{KP\mbox{-}Cov}
=
\frac{n_{\mathrm{correct}}+n_{\mathrm{partial}}}
{n_{\mathrm{all\ keypoints}}}.
\label{eq:supp-kp-cov}
\end{equation}

\noindent\textbf{Video QA (VQA) and Video QA Coverage (VQA-Cov).}
QA items are labeled as yes, partial, or no. VQA assigns scores of 1, 0.5, and 0, whereas VQA-Cov counts both yes and partial as covered:
\begin{equation}
\begin{aligned}
\mathrm{VQA}
&=\frac{n_{\mathrm{yes}}+0.5n_{\mathrm{partial}}}
{n_{\mathrm{all\ QA}}},\\
\mathrm{VQA\mbox{-}Cov}
&=\frac{n_{\mathrm{yes}}+n_{\mathrm{partial}}}
{n_{\mathrm{all\ QA}}}.
\end{aligned}
\label{eq:supp-vqa}
\end{equation}

\noindent\textbf{Reference-Tag Precision and Recall (Ref-Tag-P/R).}
These metrics evaluate reference selection using unique tags. Per-sample precision and recall are
\begin{equation}
P_i=\frac{|C_i\cap G_i|}{|C_i|},
\qquad
R_i=\frac{|C_i\cap G_i|}{|G_i|},
\label{eq:supp-reference-selection}
\end{equation}
and the reported scores are macro-averaged across samples. If a caption uses no image tag, its precision is set to zero.

\noindent\textbf{Reference Binding (Ref-Bind).}
Each locally extracted phrase--tag pair receives a continuous judge score in $[0,1]$, indicating whether the phrase's core visual referent matches the semantic description of the referenced image. Pair scores are averaged within each sample and then macro-averaged over samples.

\noindent\textbf{Effective Binding (Eff-Bind).}
Effective binding jointly reflects whether positive references are selected and whether their local phrases are correct:
\begin{equation}
\mathrm{Eff\mbox{-}Bind}
=
\mathrm{Ref\mbox{-}Tag\mbox{-}R}
\times
\mathrm{Ref\mbox{-}Bind}.
\label{eq:supp-effective-binding}
\end{equation}

\noindent\textbf{Distractor Rejection (Dist-Rej).}
For each sample,
\begin{equation}
D_i
=
1-\frac{|C_i\cap N_i|}{|N_i|}.
\label{eq:supp-distractor-rejection}
\end{equation}
The score is macro-averaged, with $D_i=1$ for samples without distractors.

\noindent\textbf{False Reference Any (FalseRef-Any).}
This metric is the fraction of distractor-containing samples in which at least one distractor tag is used:
\begin{equation}
\mathrm{FalseRef\mbox{-}Any}
=
\frac{1}{|\mathcal{I}_{N}|}
\sum_{i\in\mathcal{I}_{N}}
\mathbf{1}[C_i\cap N_i\neq\varnothing].
\label{eq:supp-false-reference-any}
\end{equation}
Lower values are better.

\noindent\textbf{Subject Recall and F1 (Subj-R/F1).}
Subject consistency is evaluated over annotated same-entity groups containing at least two positive tags. For every eligible group, extracted tags are compared with the hidden expected tags. Counts are pooled across groups:
\begin{equation}
\mathrm{Subj\mbox{-}R}
=
\frac{\sum n_{\mathrm{correct}}}{\sum n_{\mathrm{expected}}},
\qquad
\mathrm{Subj\mbox{-}P}
=
\frac{\sum n_{\mathrm{correct}}}{\sum n_{\mathrm{predicted}}},
\label{eq:supp-subject-pr}
\end{equation}
\begin{equation}
\mathrm{Subj\mbox{-}F1}
=
\frac{2\,\mathrm{Subj\mbox{-}P}\,\mathrm{Subj\mbox{-}R}}
{\mathrm{Subj\mbox{-}P}+\mathrm{Subj\mbox{-}R}}.
\label{eq:supp-subject-f1}
\end{equation}

\subsection{Evaluation Protocol}

All models receive the same task instruction, ordered reference images, and video. Content metrics operate on captions after image tags are removed. Reference-selection metrics are computed by exact matching of unique tags. Gemini-3.1-Pro is used as the MLLM judge for structured text-only evaluation of key-point coverage, video QA, reference binding, and subject consistency; final scores are computed programmatically from its structured outputs. Key-point and QA metrics use pooled item-level counts, reference-selection and binding metrics use sample-level macro averages, and subject-consistency metrics pool counts over eligible multi-reference groups.

\subsection{Qualitative Example}
\label{sec:supp-mrvbench-qualitative}

\begin{figure*}[t]
  \centering
  \includegraphics[width=\textwidth]{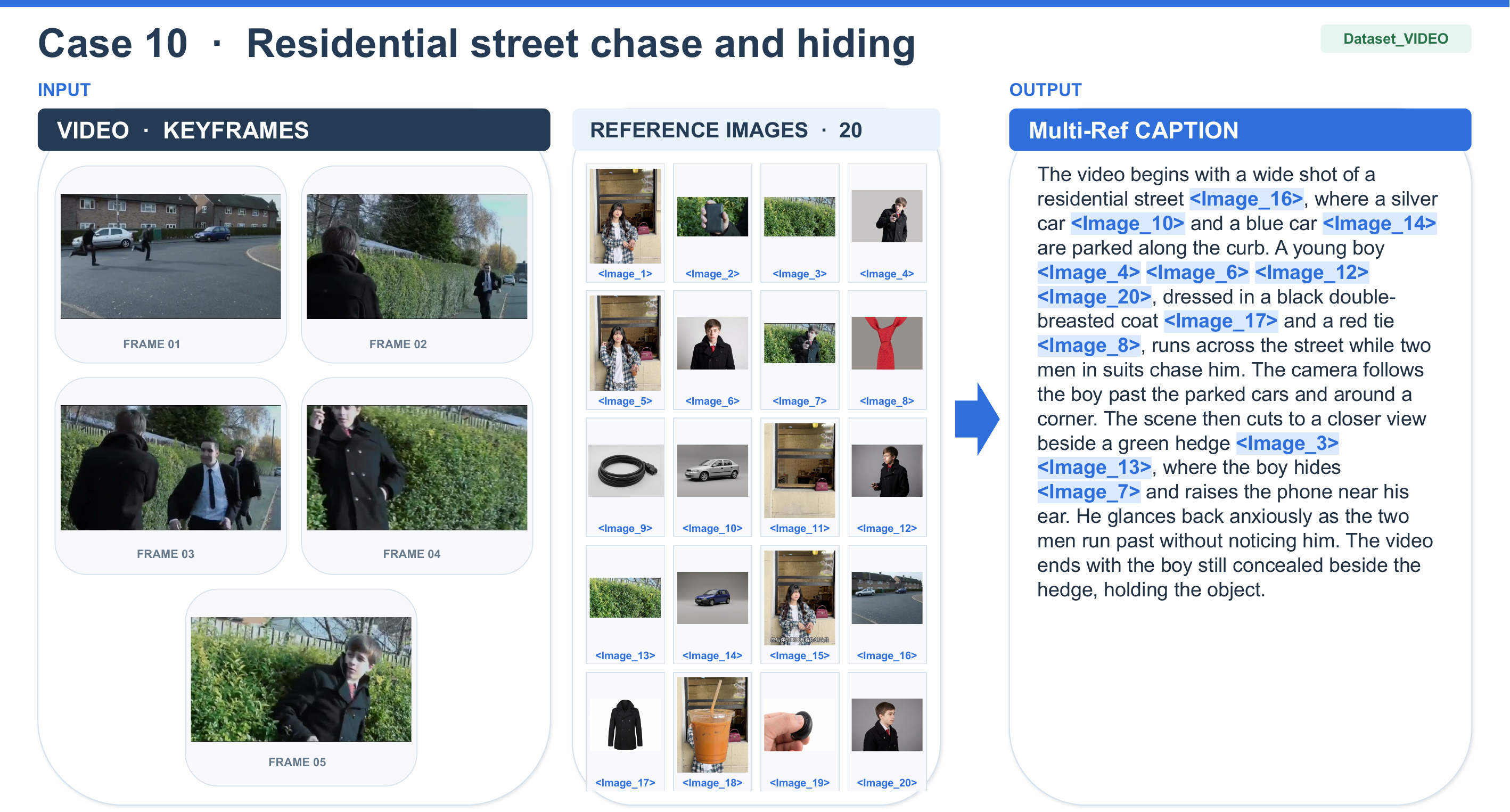}
  \caption{Qualitative example of multi-reference image-grounded video captioning on MRVBench. Given video keyframes and an ordered pool of candidate reference images, RefCaptioner generates a detailed caption with phrase-level image tags while excluding unused candidates.}
  \label{fig:supp-grounded-caption-example}
\end{figure*}

Figure~\ref{fig:supp-grounded-caption-example} shows a complete MRVBench example with video keyframes, 20 candidate reference images, and the grounded caption generated by RefCaptioner. The output places one or more image tags immediately after the corresponding local visual phrase, groups multiple references depicting the same subject or scene element, and leaves unsupported candidates unused.

\section{Model and Post-Training Details}
\label{sec:supp-training}

\subsection{Base Model and Supervised Fine-Tuning}

We use Qwen3-VL-8B-Instruct as the base vision--language model. The model receives one video together with an ordered set of tagged reference images and generates a single reference-grounded English caption. During supervised fine-tuning (SFT), the vision encoder and multimodal projector are frozen, while LoRA adapters are applied to the language model's attention and feed-forward projections. We use a LoRA rank of 64, an alpha of 128, and a dropout rate of 0.05.

SFT runs for three epochs with BF16 and DeepSpeed ZeRO-2 on $8\times$ NVIDIA H800 (80 GB) GPUs. Its environment uses Python 3.11.15, PyTorch 2.6.0+cu118, Transformers 5.8.1, and DeepSpeed 0.19.1. Per-device batch size and gradient accumulation are both 1. We use AdamW with learning rate $1\times10^{-4}$, weight decay 0.1, cosine decay, warm-up ratio 0.03, and gradient clipping 1.0. The maximum multimodal sequence length is 18,000 tokens. We validate once per epoch and select the SFT checkpoint by validation loss.

\subsection{GRPO Post-Training}
\label{sec:supp-grpo}

GRPO starts from the SFT checkpoint. The visual encoder remains frozen, while the language model is optimized without LoRA. The reward-ready training set contains 12,838 examples, and a fixed set of 126 examples is used for validation. Samples longer than 16,000 processed prompt tokens are excluded before training.

Training uses $32\times$ NVIDIA H800 (80 GB) GPUs across four nodes. The \texttt{grpo\_qwen3vl} environment uses Python 3.10.20, PyTorch 2.9.0 with CUDA 12.8, Transformers 4.57.6, vLLM 0.11.2, Ray 2.55.1, and FlashAttention 2.8.3.post1. Each update samples six responses for 16 prompts (96 captions), capped at 512 tokens each. The actor learning rate is $1\times10^{-6}$; low-variance KL regularization uses coefficient 0.005. Rollout tensor parallelism is 2, and the final policy is selected by validation performance.

\subsection{Reward Configuration}
\label{sec:supp-reward-configuration}

The GRPO reward combines factual-caption quality and multi-reference grounding quality. The caption branch rewards key-point coverage while discounting video-QA errors. The reference branch rewards correctly bound positive-reference coverage while penalizing incorrect local binding, distractor use, and inconsistent grouping of multiple references depicting the same entity.

For the multi-reference grounding reward in Eq.~\eqref{eq:reference-reward}, we use
\begin{equation}
\lambda_b=0.3,
\qquad
\lambda_d=0.4,
\qquad
\lambda_c=0.3.
\label{eq:supp-reward-coefficients}
\end{equation}
The caption-reward error-discount coefficient is $\lambda_{\mathrm{qa}}=0.5$. We combine the caption and reference branches with equal weights:
\begin{equation}
R_{\mathrm{raw}}
=0.5R_{\mathrm{cap}}+0.5R_{\mathrm{ref}}.
\label{eq:supp-raw-reward}
\end{equation}
Without a structural violation, $R=R_{\mathrm{raw}}$. For malformed, invented, non-local, or prohibited tags, we use $\tau_{\mathrm{struct}}=0.2$ and cap the reward as
\begin{equation}
R=\min\!\left(R_{\mathrm{raw}},\tau_{\mathrm{struct}}\right)
=\min\!\left(R_{\mathrm{raw}},0.2\right).
\label{eq:supp-structural-cap}
\end{equation}

\subsection{Visual Preprocessing}

For SFT, videos are sampled at a target rate of 2 fps. Each video frame is limited to 602,112 pixels, and each reference image is limited to 401,408 pixels. Samples exceeding the sequence-length limit after multimodal processing are removed rather than truncated. GRPO uses the same frame-rate and pixel constraints.

\subsection{Repeated Runs and Statistical Significance}

We use a fixed construction seed of 20260710. The key comparison between the base model and the fully post-trained RefCaptioner is independently repeated three times. Across these repeated runs, the post-trained model consistently outperforms the base model, and the improvement is statistically significant.

\section{General Video-Captioning Evaluation}
\label{sec:supp-general-evaluation}

\subsection{Evaluation Setup}

All baseline models are evaluated using the same video inputs and benchmark-specific instructions. We use deterministic decoding for caption generation. The benchmark-specific frame-sampling rates, generation lengths, and evaluation procedures are described below.

\subsection{VDC}

For VDC, each model is prompted to provide a faithful, detailed description of the video in more than three sentences. Videos are sampled at 1 fps with a maximum of eight frames. Caption generation permits up to 2,048 new tokens.

Evaluation follows VDC's two-stage question-based protocol. For each ground-truth question, a text judge first answers the question using only the generated caption. A second judgment compares this answer with the reference answer and returns a semantic-match decision together with a score from 0 to 5. Scores and match accuracies are averaged over questions and samples. We use Gemini-3.1-Pro as the MLLM judge while preserving the benchmark's question--answer and aggregation procedure. Evaluation covers 1,027 videos.

\subsection{VCapsBench}

For VCapsBench, we use the benchmark instruction requesting a comprehensive caption of at least 200 words, covering actions, characters, spatial relations, objects, emotions, visual techniques, background, lighting and color, shots, and overall style. Videos are sampled at 2 fps with a maximum of 64 frames. Caption generation permits up to 1,024 new tokens.

The generated caption is evaluated against the benchmark's video-specific QA bank. For each QA item, a text judge outputs \texttt{Yes}, \texttt{No}, or \texttt{Unanswerable}. We report answer rate, inaccuracy rate, and coverage rate:
\begin{equation}
\begin{aligned}
\mathrm{AnswerRate}
&=\frac{N_{\mathrm{correct}}}{N_{\mathrm{total}}},\\
\mathrm{InaccuracyRate}
&=\frac{N_{\mathrm{incorrect}}}
{N_{\mathrm{total}}-N_{\mathrm{unanswerable}}},
\end{aligned}
\label{eq:supp-vcaps-accuracy}
\end{equation}
\begin{equation}
\mathrm{CoverageRate}
=
1-\frac{N_{\mathrm{unanswerable}}}{N_{\mathrm{total}}}.
\label{eq:supp-vcaps-coverage}
\end{equation}
We use Gemini-3.1-Pro as the MLLM judge and retain the benchmark's answer mapping and aggregation. Throughout evaluation, the same judging configuration is applied to all captioning methods so that every prediction is processed under a consistent protocol. The judge receives the benchmark-defined inputs and produces outputs in the expected response format, after which the original answer mapping is used without modification. We then aggregate the resulting per-sample decisions according to the official benchmark procedure. This standardized pipeline avoids method-specific post-processing and ensures that comparisons are computed on a common basis. Evaluation is performed over all 5,666 unique videos in our evaluation set, providing broad benchmark coverage while keeping the scoring procedure identical for every model.

\section{Prompt and Input Formats}
\label{sec:supp-prompts}

\subsection{Training Input}

Each sample contains an ordered list of reference images followed by one video. Reference images are paired positionally with tags \texttt{<Image\_1>}, \texttt{<Image\_2>}, and so forth. Some references may be distractors and therefore need not appear in the output. The model receives a single user instruction, and the target caption is used as the assistant response.

The multimodal content is serialized as follows:

\begin{tcolorbox}[suppprompt,colback=suppSoftGreen,colframe=suppGreen,
  title=Multimodal Input Serialization]
\begin{Verbatim}[breaklines=true,breakanywhere=true,fontsize=\small]
Reference image tags, in order: <Image_1>, ..., <Image_N>

<Image_1> reference image: [reference image 1]
...
<Image_N> reference image: [reference image N]

Reference video: [video]

Now write the final caption.
\end{Verbatim}
\end{tcolorbox}

The training instruction is:

\begin{tcolorbox}[suppprompt,colback=suppSoftBlue,colframe=suppBlue,
  title=Training Instruction]
\begin{Verbatim}[breaklines=true,breakanywhere=true,fontsize=\small]
You are a multi-reference video captioning model.

Input:
- Several reference images, each labeled as <Image_1>, <Image_2>, etc.
- Some reference images may be distractors that are not visible in the video.
- One reference video.

Task:
Write one fluent English video caption that describes the visible video content and locally binds each usable reference image tag to the visual phrase it grounds.

Return only one natural English paragraph. Describe the visible subjects, setting, referenced appearances or objects, action progression, and useful camera or visual-style details. Do not invent unseen names, relationships, causes, dialogue, audio, or story details.

Use only provided tags that can be confidently grounded in the video. Place each used tag immediately after a concrete noun phrase. Stack multiple tags only when they describe the same visual unit; otherwise attach them to separate phrases. Do not attach tags to pronouns, collect tags at the end, or use phrases such as "from <Image_N>" or "shown in <Image_N>". Omit ungrounded distractor tags.
\end{Verbatim}
\end{tcolorbox}

An example output is:

\begin{tcolorbox}[suppprompt,colback=suppSoftYellow,colframe=suppOrange,
  title=Example Target Caption]
\begin{Verbatim}[breaklines=true,breakanywhere=true,fontsize=\small]
The scene opens in a brightly lit laboratory, where a young woman <Image_3> <Image_5> works at a desk. She wears a white lab coat <Image_7> and clear safety glasses <Image_2>, while a microscope <Image_4> sits beside her. She studies her computer, briefly raises a hand to her chin, and then returns her attention to the screen as the camera slowly pans right.
\end{Verbatim}
\end{tcolorbox}

The two tags after ``young woman'' form one same-entity group, whereas the lab coat, safety glasses, and microscope are bound to separate local phrases. Any reference not visibly grounded in the video is omitted.

\subsection{Evaluation Prompts}

We use separate text-only judge prompts for content coverage and multi-reference grounding. Tags are removed before content evaluation but retained for grounding evaluation.

\noindent\textbf{Key-point coverage.}
The judge receives a candidate caption and a frozen list of golden key points. It labels every key point as \texttt{correct}, \texttt{partial}, or \texttt{missing}. Paraphrases are accepted; distorted, incomplete, or incorrectly attributed content is partial; unsupported content is missing. The judge returns one JSON record per key point.

\begin{tcolorbox}[suppprompt,colback=suppSoftPurple,colframe=suppPurple,
  title=Key-Point Coverage Judge Prompt]
\begin{Verbatim}[breaklines=true,breakanywhere=true,fontsize=\small]
For each golden key point, determine whether the candidate caption covers it. Use only the caption and the supplied key points. Judge each item independently and return JSON labels: correct, partial, or missing.
\end{Verbatim}
\end{tcolorbox}

The key-point bank covers six complementary dimensions, summarized in Table~\ref{tab:supp-keypoint-bank}.

\begin{table*}[t]
  \centering
  \begingroup
  \small
  \setlength{\tabcolsep}{4pt}
  \renewcommand{\arraystretch}{1.15}
  \begin{tabular}{p{0.12\textwidth}p{0.34\textwidth}p{0.46\textwidth}}
    \toprule
    \textbf{Dimension} & \textbf{Definition} & \textbf{Representative golden key point} \\
    \midrule
    Subject & Principal visible entities or objects, including identity, role, or count. & The video features a female character with a white robotic body. \\
    Appearance & Visible clothing, color, material, body features, accessories, or expression. & The character wears a tan poncho with brown geometric patterns. \\
    Action & Actions, interactions, state changes, and temporal event sequences. & The character pushes the yellow chair from left to right across the snowy ground. \\
    Background & Location, surrounding objects, weather, lighting, and spatial context. & A green car covered in snow is parked near a decorated house in the background. \\
    Camera & Shot scale, viewpoint, angle, camera motion, reframing, or cuts. & The camera follows the character, panning slightly to the right. \\
    Style & Rendering mode, palette, cinematic treatment, mood, or atmosphere. & The scene has a festive, wintery atmosphere. \\
    \bottomrule
  \end{tabular}
  \endgroup
  \caption{Definitions and representative examples for the six key-point dimensions.}
  \label{tab:supp-keypoint-bank}
\end{table*}

For the action example, ``She pushes a yellow chair from left to right through the snow'' is \texttt{correct} because it preserves the actor, object, direction, and setting. ``She pushes a chair across the snow'' is \texttt{partial} because the color and direction are omitted. ``She stands in a festive snowy village'' is \texttt{missing} because it does not describe the chair-pushing action.

\noindent\textbf{Video QA.}
The judge receives the candidate caption and frozen question--answer pairs. For each item, it determines whether the caption supports the reference answer and returns \texttt{yes}, \texttt{partial}, or \texttt{no}. Semantic equivalence is accepted, but unsupported inference is prohibited.

\begin{tcolorbox}[suppprompt,colback=suppSoftTeal,colframe=suppTeal,
  title=Video-QA Judge Prompt]
\begin{Verbatim}[breaklines=true,breakanywhere=true,fontsize=\small]
For each video-content QA item, determine whether the candidate caption supports the reference answer. Use only the caption and the supplied QA items, and return one JSON label for every item: yes, partial, or no.
\end{Verbatim}
\end{tcolorbox}

The six QA types are summarized in Table~\ref{tab:supp-qa-bank}.

\begin{table*}[t]
  \centering
  \begingroup
  \small
  \setlength{\tabcolsep}{4pt}
  \renewcommand{\arraystretch}{1.15}
  \begin{tabular}{p{0.18\textwidth}p{0.32\textwidth}p{0.42\textwidth}}
    \toprule
    \textbf{QA type} & \textbf{Representative question} & \textbf{Reference answer} \\
    \midrule
    Entity action & What sequence of actions does the young man perform? & He looks down, sits on a ledge and drinks, looks over the skyline, then lowers the bottle and rests his hands on his lap. \\
    Entity background & What is parked near the decorated house? & A green car covered in snow. \\
    Multi-entity interaction & What is the woman doing with the man whose back faces the camera? & She is struggling with him. \\
    Camera action & How does the camera frame the man throughout the video? & It begins at a low angle, cuts to a side profile, and then shows him from behind. \\
    Appearance entity & What is the female character wearing? & A tan hat with a brown band and a tan poncho with brown geometric patterns. \\
    Style scene & What is the visual style of the scene? & A slightly muted palette gives the scene a gritty appearance. \\
    \bottomrule
  \end{tabular}
  \endgroup
  \caption{Question types and representative examples in the video-QA bank.}
  \label{tab:supp-qa-bank}
\end{table*}

For the camera-action example, ``The view moves from a low angle to his side profile and finally to a rear view'' is labeled \texttt{yes}. ``The camera cuts to his side profile'' is \texttt{partial} because it omits two viewpoints. ``He drinks while looking toward the skyline'' is \texttt{no} because it describes subject action rather than camera behavior.

\noindent\textbf{Reference binding.}
For every locally extracted phrase--tag pair, the judge compares the phrase's core visual referent with the semantic description of that reference image. It scores the pair from 0 to 1. Synonyms, compatible visual categories, and changes in pose or temporal context are allowed; wrong referents, conflicting attributes, vague phrases, and unsupported bindings are penalized.

\begin{tcolorbox}[suppprompt,colback=suppSoftPink,colframe=suppPink,
  title=Reference-Binding Judge Prompt]
\begin{Verbatim}[breaklines=true,breakanywhere=true,fontsize=\small]
For each phrase-tag pair, identify the core visual referent and determine whether it belongs to the semantic description of that reference image. Return a score in [0,1] and a short reason.
\end{Verbatim}
\end{tcolorbox}

For example, if \texttt{<Image\_7>} and \texttt{<Image\_8>} both depict a black velvet dress with gold embroidery, the local phrase ``a black velvet dress with gold embroidery \texttt{<Image\_7>} \texttt{<Image\_8>}'' receives a high binding score. Binding \texttt{<Image\_7>} to ``a wide-brimmed hat'' is incorrect because the core visual referent does not match that reference description.

\noindent\textbf{Subject consistency.}
The judge acts as a reference-group extractor. Given a candidate caption and semantic same-entity groups, it assigns locally attached candidate tags to each group without seeing the expected tags. The extracted assignments are then compared programmatically with the hidden group annotations.

\begin{tcolorbox}[suppprompt,colback=suppSoftOrange,colframe=suppDeepOrange,
  title=Subject-Consistency Judge Prompt]
\begin{Verbatim}[breaklines=true,breakanywhere=true,fontsize=\small]
Map tags appearing in the candidate caption to the same-entity reference group they locally describe. Use only local phrase-tag evidence. Return the predicted tag set for every group in JSON; return an empty set when the group is absent or untagged.
\end{Verbatim}
\end{tcolorbox}

Consider a dress group whose hidden expected set is $E=\{\texttt{<Image\_7>},\texttt{<Image\_8>}\}$. The fragment ``a black velvet dress with gold embroidery \texttt{<Image\_7>} \texttt{<Image\_8>}'' yields $P=E$ and is fully consistent. Omitting \texttt{<Image\_8>} creates a missing-tag error and lowers recall. Adding an unrelated hat tag to the dress phrase creates an intruder error and lowers precision and Subj-F1.

\section{Downstream Video Reconstruction}
\label{sec:supp-reconstruction}

\subsection{Reconstruction Protocol}
\label{sec:supp-reconstruction-protocol}

We evaluate whether reference-grounded captioner outputs provide effective conditioning for downstream video reconstruction. For the controlled Wan 2.1 VACE experiment, we use Wan2.1-VACE-14B. For each captioner, only the reference images explicitly cited by its generated caption are provided to the generator, and all \texttt{<Image\_N>} tokens are removed from the text before generation.

Each output contains 81 frames. Landscape videos are generated at $1280\times720$, and portrait videos at $720\times1280$. We use seed 0, 50 sampling steps, and a guidance scale of 5.0 for all caption sources. The reconstruction settings are therefore fixed, while each captioner supplies its generated description and selected reference-image subset. This setup evaluates the end-to-end utility of caption generation and reference selection for preserving subjects, referenced attributes, actions, and scene content during reconstruction.

\subsection{Qualitative Comparisons}
\label{sec:supp-reconstruction-comparisons}

\begin{figure*}[t]
  \centering
  \includegraphics[width=\textwidth]{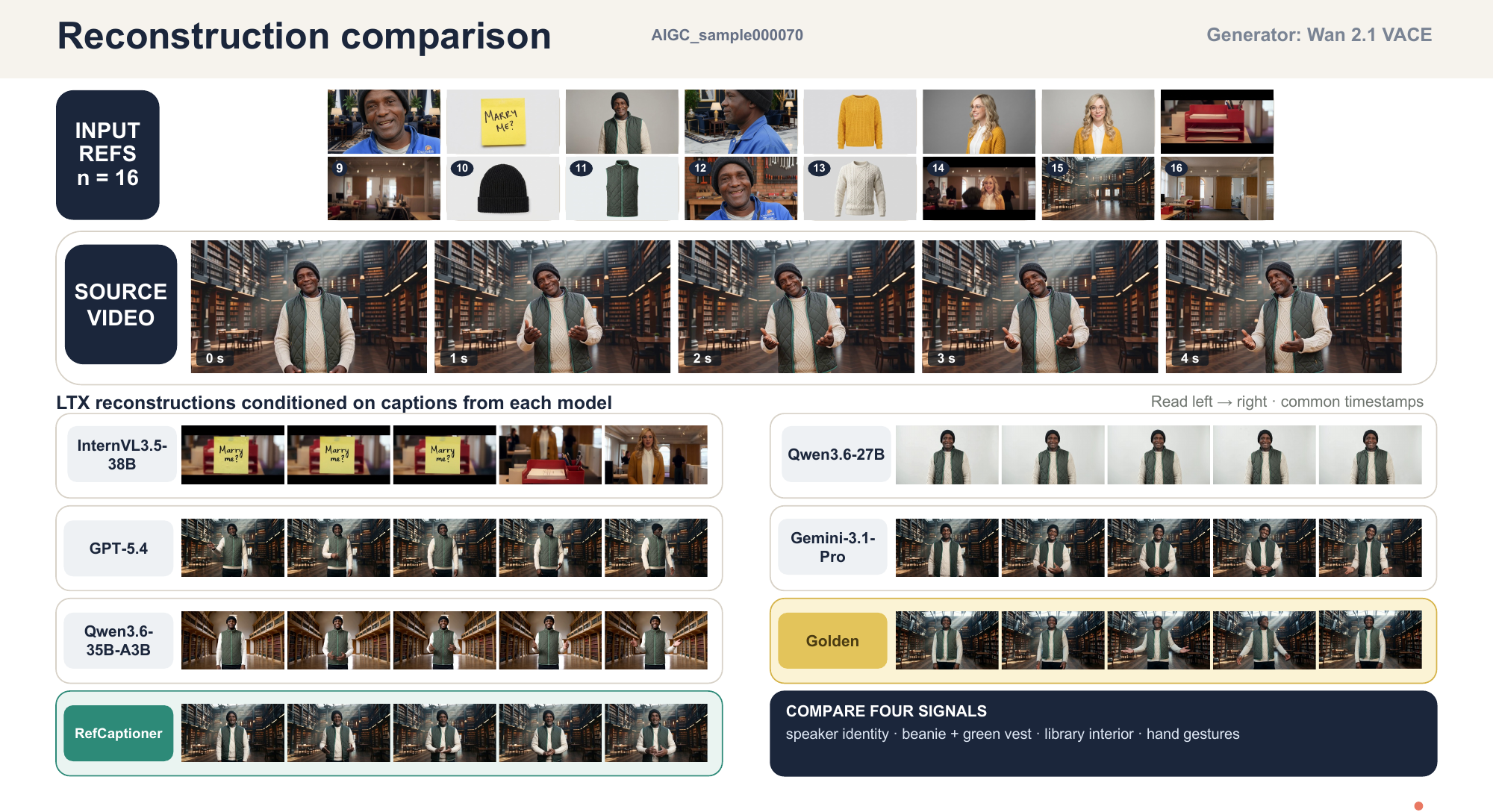}
  \caption{Caption-conditioned reconstruction comparison on AIGC sample 000070.
  The figure compares common timestamps from the source video and Wan 2.1 VACE reconstructions conditioned on the caption and selected reference-image subset produced by each captioner.}
  \label{fig:supp-casecaption-library}
\end{figure*}

\begin{figure*}[p]
  \centering
  \includegraphics[width=\textwidth]{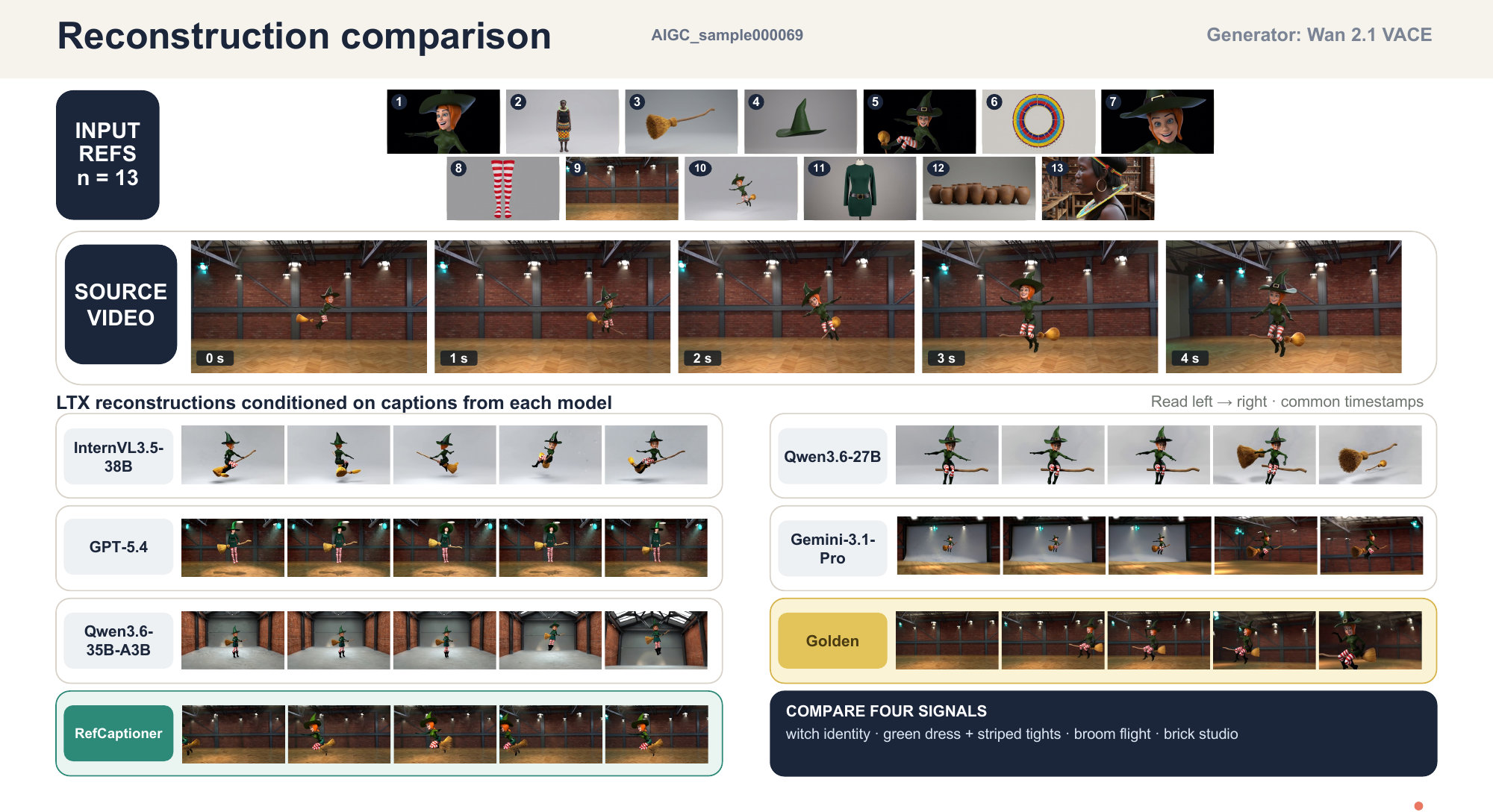}

  \includegraphics[width=\textwidth]{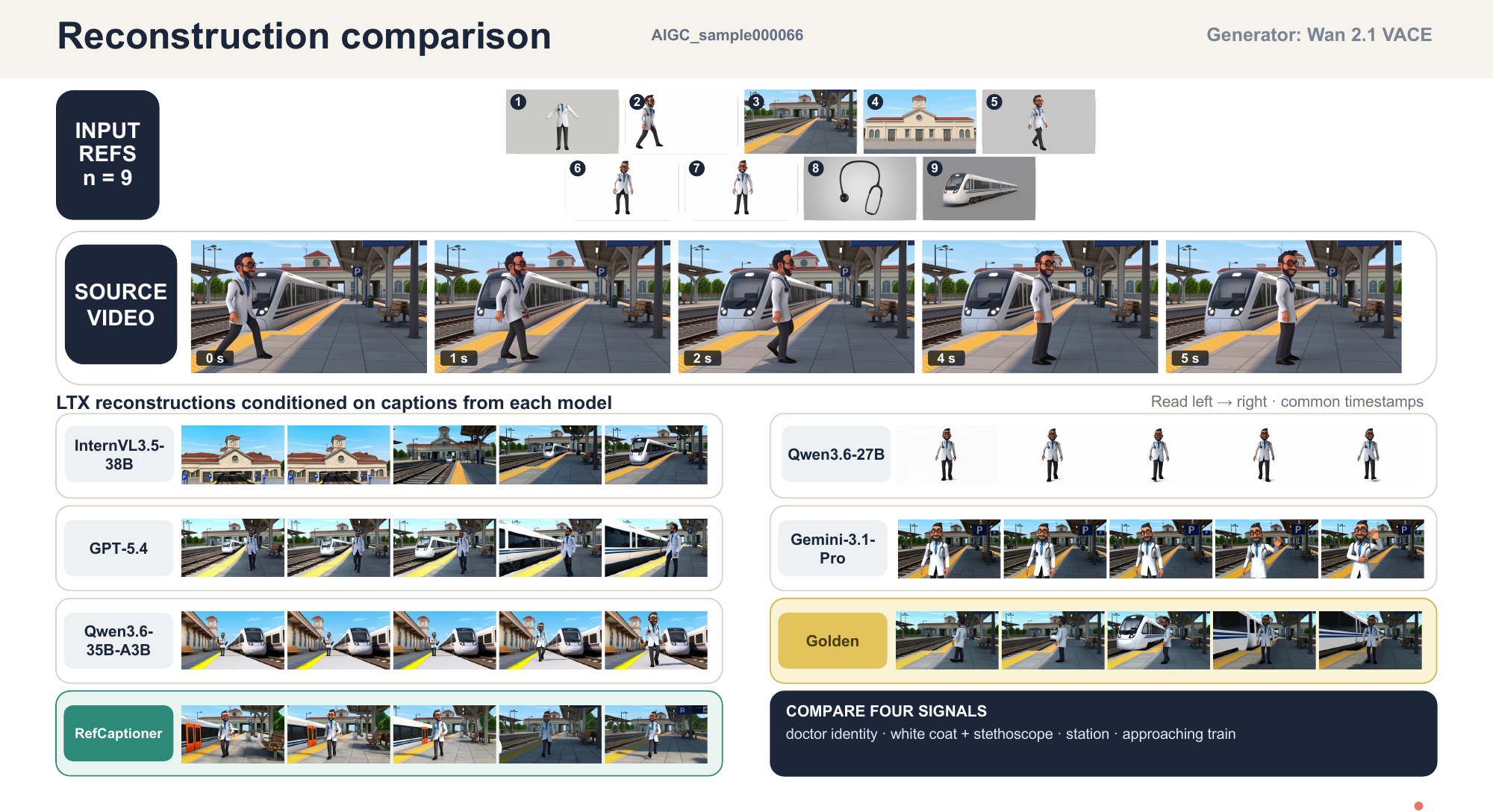}
  \caption{Additional caption-conditioned reconstruction comparisons using Wan 2.1 VACE.
  Results are shown for AIGC samples 000069 (top) and 000066 (bottom). All methods use the same generation settings, while each reconstruction uses the caption and selected reference-image subset produced by the corresponding captioner.}
  \label{fig:supp-casecaption-wan}
\end{figure*}

\begin{figure*}[p]
  \centering
  \includegraphics[width=\textwidth]{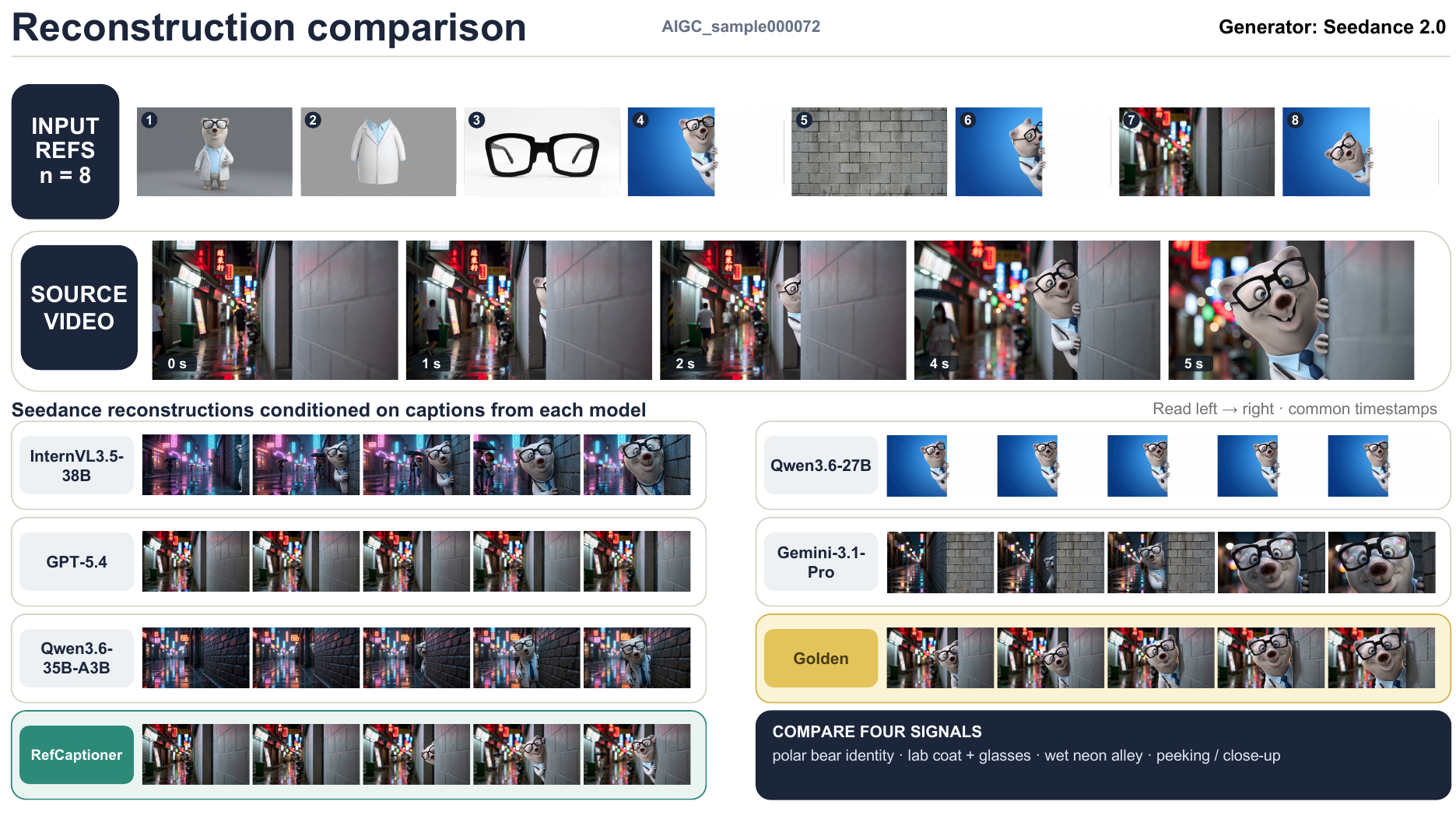}

  \includegraphics[width=\textwidth]{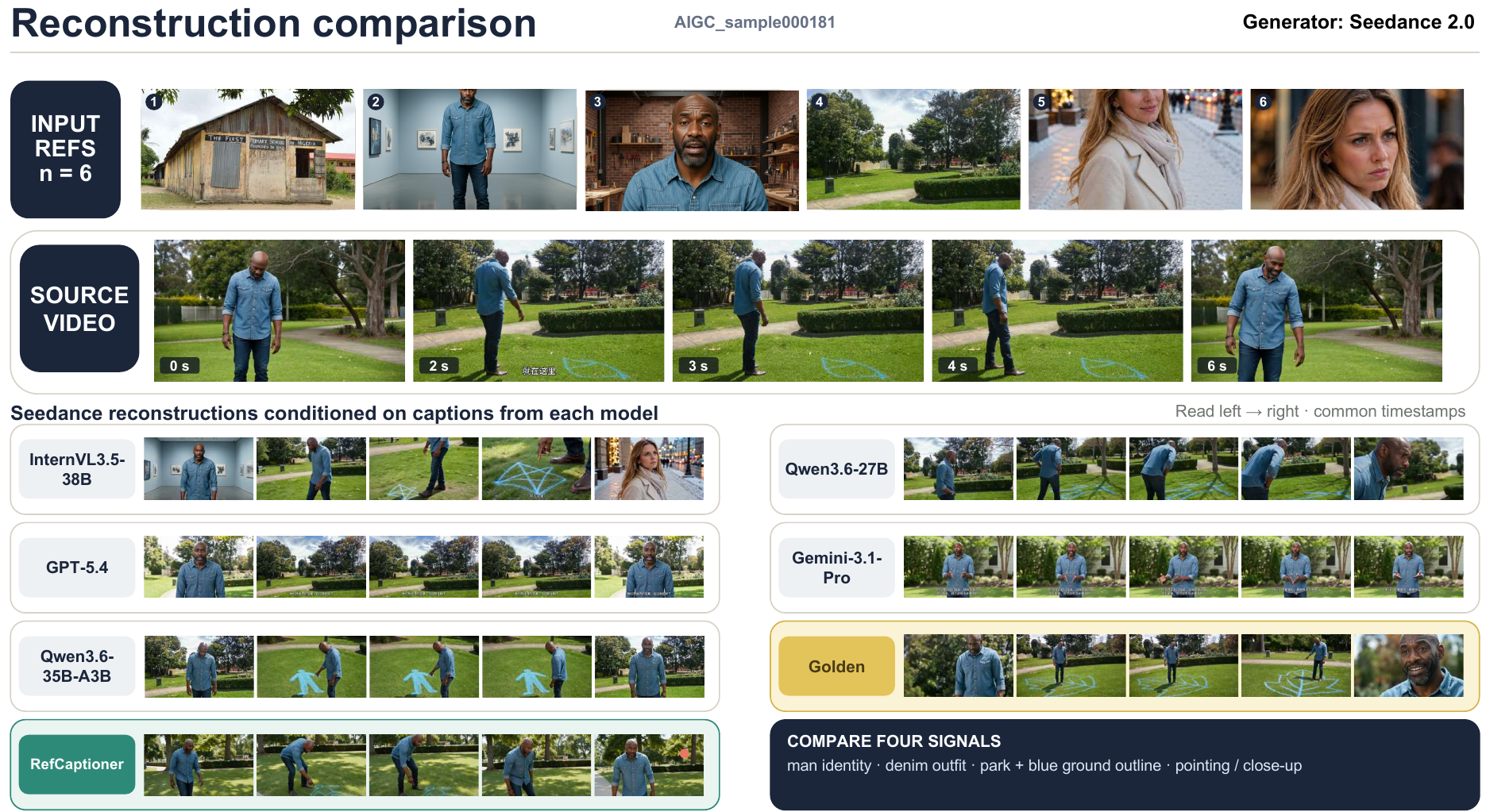}
  \caption{Additional caption-conditioned reconstruction comparisons using Seedance 2.0.
  Results are shown for AIGC samples 000072 (top) and 000181 (bottom). All methods use the same generation settings, while each reconstruction uses the caption and selected reference-image subset produced by the corresponding captioner.}
  \label{fig:supp-casecaption-seedance}
\end{figure*}

Figures~\ref{fig:supp-casecaption-library}--\ref{fig:supp-casecaption-seedance} provide qualitative comparisons using Wan 2.1 VACE and Seedance 2.0. Within each reconstruction backend, all captioners use the same generation settings, while each reconstruction is conditioned on the caption and reference-image subset produced by the corresponding captioner.

\section{Human Evaluation Details}
\label{sec:supp-human-evaluation}

\subsection{Human Preference Study}
\label{sec:human-preference-study}

We recruited three graduate-level researchers with relevant expertise as human evaluators. We randomly selected 40 examples from MRVBench, comprising 20 AIGC videos and 20 real videos. For each example, the evaluators viewed the source video, the candidate reference images, and captions generated by GPT-5.4, Gemini-3.1-Pro, Qwen3.6-35B-A3B, Qwen3-VL-8B-Instruct, and RefCaptioner. Model identities were hidden, and the presentation order was independently randomized for each sample to reduce model-identity and position bias.

Each evaluator independently produced a complete ranking of the five captions according to overall preference and naturalness, with rank 1 denoting the most preferred output and rank 5 the least preferred. Mean rank for each model was computed across all 40 samples and all three evaluators; lower values indicate stronger human preference. Separately from the preference ranking, the evaluators judged whether each reference image was bound to the correct local phrase in each caption. Binding accuracy was computed by aggregating these independent reference-level judgments across samples and evaluators.

We recruited graduate-student annotators and randomly selected 40 examples from MRVBench, comprising 20 AIGC videos and 20 real videos. For each example, participants viewed the video, the reference images, and anonymized captions from GPT-5.4, Gemini-3.1-Pro, Qwen3.6-35B-A3B, Qwen3-VL-8B-Instruct, and RefCaptioner. The caption order was randomized and model identities were hidden. Participants ranked the five captions by overall preference and naturalness, where a lower mean rank indicates a better result, and separately judged whether each reference image was bound to the correct phrase. As shown in Fig.~\ref{fig:human-preference-study}, RefCaptioner achieved the best mean rank of 1.75 and a binding accuracy of 0.99. Its binding accuracy was close to the 1.00 achieved by Gemini-3.1-Pro and GPT-5.4, and exceeded the 0.96 and 0.93 achieved by Qwen3.6-35B-A3B and Qwen3-VL-8B-Instruct. Although all models obtained high binding accuracy, RefCaptioner received a clearly better preference rank. Correct image insertion alone therefore does not determine human preference; caption structure and expression also matter. We organized the training captions around six components---subjects, appearance, actions, background, visual style, and camera work---and trained the model to bind each component to the appropriate reference rather than repeatedly attach one image to several phrases. As a result, RefCaptioner produces accurately grounded captions that more closely match how people organize and express video descriptions.

\begin{center}
  \begin{minipage}{\columnwidth}
  \centering
  \includegraphics[width=\linewidth]{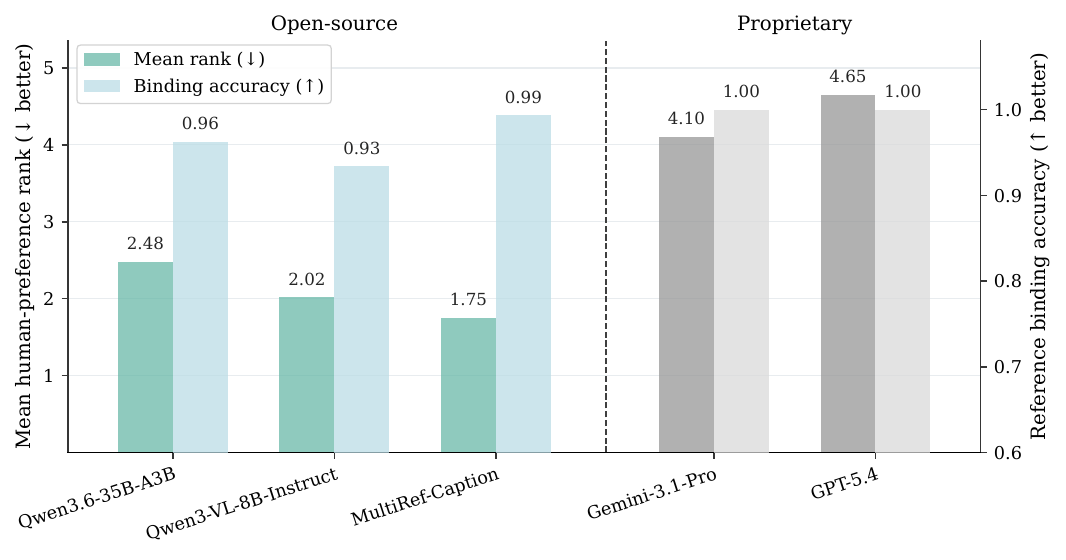}
  \captionof{figure}{\textbf{Human preference and binding accuracy.}
  Lower rank is better; MultiRef-Caption denotes RefCaptioner.}
  \label{fig:human-preference-study}
  \end{minipage}
\end{center}

\clearpage

\subsection{Reconstruction Evaluation}

The same three experts evaluated the caption-conditioned video reconstructions through pairwise Good/Same/Bad (GSB) comparisons. For each evaluated source video and each baseline captioner, the experts viewed the source video together with two anonymized reconstructions: one conditioned on the output of RefCaptioner and the other on the output of the baseline. The left--right presentation order was independently randomized for each comparison, and model identities were hidden. Each expert assigned \emph{Good} when the RefCaptioner-conditioned reconstruction agreed better with the source video, \emph{Same} when the two reconstructions were comparable, and \emph{Bad} when the baseline-conditioned reconstruction agreed better. Judgments considered subjects, appearances, actions, objects, backgrounds, and event progression. The reported percentages are aggregated over all pairwise judgments.

Because each captioner supplies both a generated caption and the subset of reference images cited by that caption, this evaluation measures their combined downstream utility rather than isolating textual caption quality alone. The reconstruction backend, generation settings, and random seed are held fixed across captioners, as detailed in Appendix~\ref{sec:supp-reconstruction-protocol}.

\section{Use of AI Tools}
\label{sec:supp-ai-disclosure}

Generative AI tools were used to polish author-provided prose and to assist in writing data-preprocessing and experimental code. All AI-assisted text and code were reviewed, tested, and verified by the authors, who take full responsibility for the content of the submission.

\end{document}